%% file: main.tex
\newif\ifcvpr
\newif\ifreview
\cvprfalse
\reviewtrue

\documentclass[10pt,twocolumn,letterpaper]{article}

\ifcvpr
  \ifreview
    \usepackage[review]{cvpr}
    \PassOptionsToPackage{pagebackref}{hyperref}
  \else
    \usepackage{cvpr}
  \fi
\fi
\input{preamble}
\usepackage{balance}
\ifcvpr
  \def\paperID{*****}
  \def\confName{CVPR}
  \def\confYear{2027}
\else
  \usepackage[top=1in,textheight=8.875in,textwidth=6.875in,columnsep=0.3125in]{geometry}
  \usepackage{enumitem}
  \usepackage[font=small,skip=6pt]{caption}
  \setlist[itemize]{leftmargin=*,itemsep=2pt,topsep=3pt}
  \date{}
  \makeatletter
  \renewcommand{\@maketitle}{%
    \newpage\null
    \begin{center}
      {\Large\bfseries\@title\par}
      \vskip 14pt
      {\large\lineskip .5em\begin{tabular}[t]{c}\@author\end{tabular}\par}
      \vskip 10pt
    \end{center}%
  }
  \makeatother
  \renewenvironment{abstract}{%
    \par\centerline{\large\bfseries Abstract}\vspace{8pt}
    \noindent\itshape\ignorespaces
  }{\par\vspace{10pt}}
\fi

\title{Fight World Model: Customizable Combat through Visual Interaction Proxies}
\author{%
  Jiahua Li\textsuperscript{1}\quad
  Zixu John\textsuperscript{2}\quad
  Tom Zhong\textsuperscript{2}\quad
  Fuping Wu\textsuperscript{3}\quad
  Tianhao Xu\textsuperscript{2}\\
  [3pt]
  Ruicheng Zhang\textsuperscript{4}\quad
  Jun Zhou\textsuperscript{4}\quad
  Zunnan Xu\textsuperscript{5,}\thanks{Corresponding Author.}\\
  [3pt]
  {\small\textsuperscript{1}University of Oxford, UK\quad
  \textsuperscript{2}Independent Researcher}\quad
  {\small\textsuperscript{3}Imperial College London, UK}
  \\
  {\small\textsuperscript{4}Tsinghua University, China}\quad
  {\small\textsuperscript{5}Nanyang Technological University, Singapore}
}
\hypersetup{pdftitle={Fight World Model: Customizable Combat through Visual Interaction Proxies}}
\hypersetup{pdfauthor={Jiahua Li, Zixu John, Tom Zhong, Fuping Wu, Tianhao Xu, Ruicheng Zhang, Jun Zhou, Zunnan Xu}}
\ifcvpr
  \ifreview
    \hypersetup{pdfauthor={}}
  \fi
\fi

\makeatletter
\g@addto@macro\@maketitle{%
  \begingroup
    \centering
    \includegraphics[width=\textwidth]{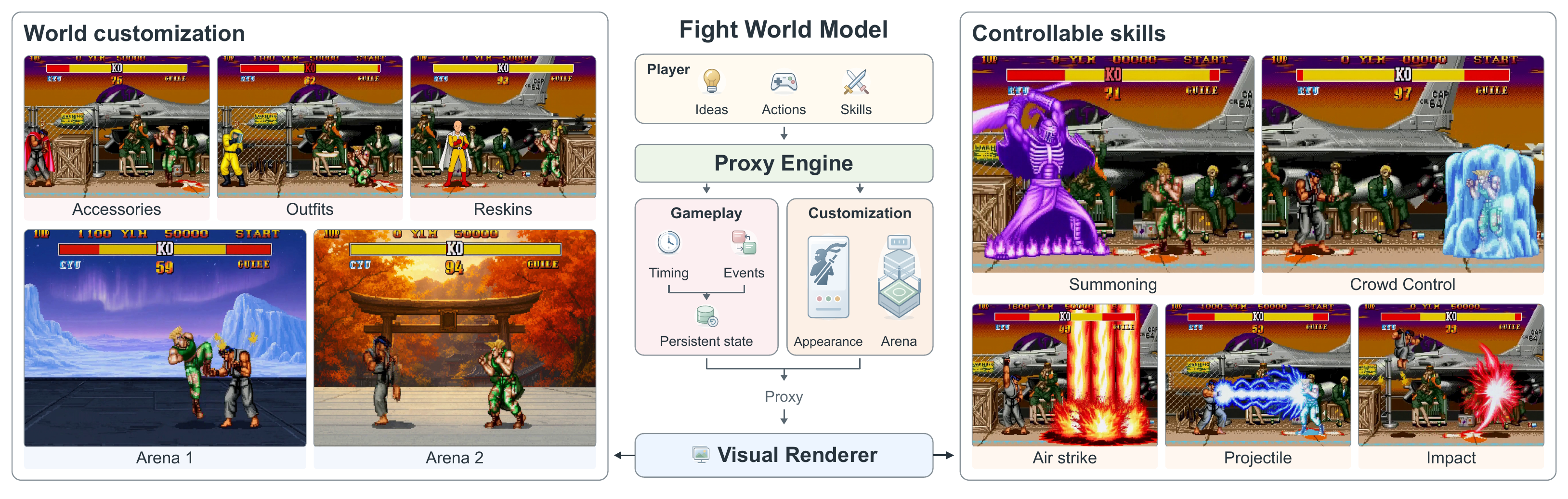}
    \def\@captype{figure}
    \caption{\textbf{Customizable Combat with FWM.} Character and arena customization is shown on the left, and controllable skills on the right. The central pipeline connects rule execution to video rendering through an interaction proxy.}
    \label{fig:motivation}
    \par
  \endgroup
  \par\vspace{8pt}
}
\makeatother

\begin{document}
\maketitle

\begin{abstract}

A playable, programmable game world must maintain persistent state, enforce executable rules, and support extensible gameplay mechanics. Existing video world models can generate plausible combat, yet struggle to keep hit reactions and health updates consistent with game rules. We introduce Fight World Model (FWM), which separates rule-governed combat execution from video rendering. An executable engine maintains state and resolves interactions. A deterministic compiler encodes these outcomes and attempted actions in a spatially and temporally aligned interaction proxy. The proxy aligns action and skill windows, brief hit reactions, and persistent health by fighter identity and timing. These cues can be controlled separately to render interactions under modified rules. Identity-bound references preserve character appearance across side switches. We train FWM on FightWorld Corpus, spanning 52 skill effects, 187 character-appearance variants, and 15 arenas. Without retraining the renderer, FWM supports damage scaling, limited-hit immunity, lifesteal, and multi-stage skill mechanisms beyond the native game. It also allows character and arena customization. Experiments demonstrate the effectiveness of FWM, which outperforms existing game world models in action fidelity, hit reactions, health-state consistency, and character appearance preservation. Our project is available at \href{https://fight-world-model.github.io}{fight-world-model.github.io}.
\end{abstract}
\section{Introduction}
\label{sec:intro}

Interactive world models aim to simulate how environments evolve under user actions~\cite{ha2018worldmodels,bruce2024genie,deepmind2025genie3,zhang2025mind,zhang2026robostereo}. Recent video world models learn from gameplay recordings to generate playable sequences from controller or text inputs~\cite{valevski2024gamengen,decart2024oasis,yu2025gamefactory,wang2026matrixgame3,robbyant2026lingbotworld,hu2026wantofight}. Customizing such worlds requires changing both gameplay rules and visual appearances. In multi-fighter combat, the same attack may cause different health loss under different rules. A protected fighter may react to a hit without losing health, while lifesteal may restore the attacker's health. When only character or arena appearance changes, the specified interactions should remain the same. As illustrated in \cref{fig:motivation}, a customizable system must allow interaction outcomes to follow edited rules while preserving those interactions across appearance changes.

These requirements are difficult to meet when state transitions are inferred within the video generator~\cite{valevski2024gamengen,decart2024oasis,wang2026matrixgame3,hu2026wantofight,chen2026h3world,wang2026reactivegwm_v1}. The model learns how actions relate to reactions and state changes from recorded gameplay. StatePlay further predicts game-state variables, such as health, alongside video frames~\cite{lin2026stateplay}, but predicting state does not itself provide an editable rule for updating it. Multi-entity action binding, identity-aware instructions, and NPC planning extend control over characters~\cite{zhu2026incantation,pondaven2026actionparty,deng2026worldmind,wang2026reactivegwm_v1,wang2026reactivegwm_v2}. They do not by themselves determine how interaction outcomes should change under a new rule configuration. When rules change, a generator may retain familiar action--consequence associations from training and deviate from the externally specified outcomes.

Code-driven world models offer a way to execute world-state transitions outside the video generator and render the resulting spatial proxies~\cite{chen2026cwm,huang2026pwm,zhan2026magpie,yang2026oneira}. Correct state execution alone, however, does not ensure that the generated interaction follows those results. The renderer must identify the acting and affected fighters, depict each event at the intended time, and preserve the resulting state across subsequent exchanges. Spatial layouts alone do not specify action windows, hit reactions, or persistent health changes. Text and state tokens can carry these quantities, but the renderer must map them to the appropriate body regions, heads-up display (HUD) elements, and video frames. Appearance customization adds a related difficulty: actions and consequences must remain associated with the intended fighters as their appearances change or they exchange positions. The interface must therefore convey attempted actions, events, and persistent state with consistent identity and timing.

\begin{figure*}[!t]
  \centering
  \includegraphics[width=\textwidth]{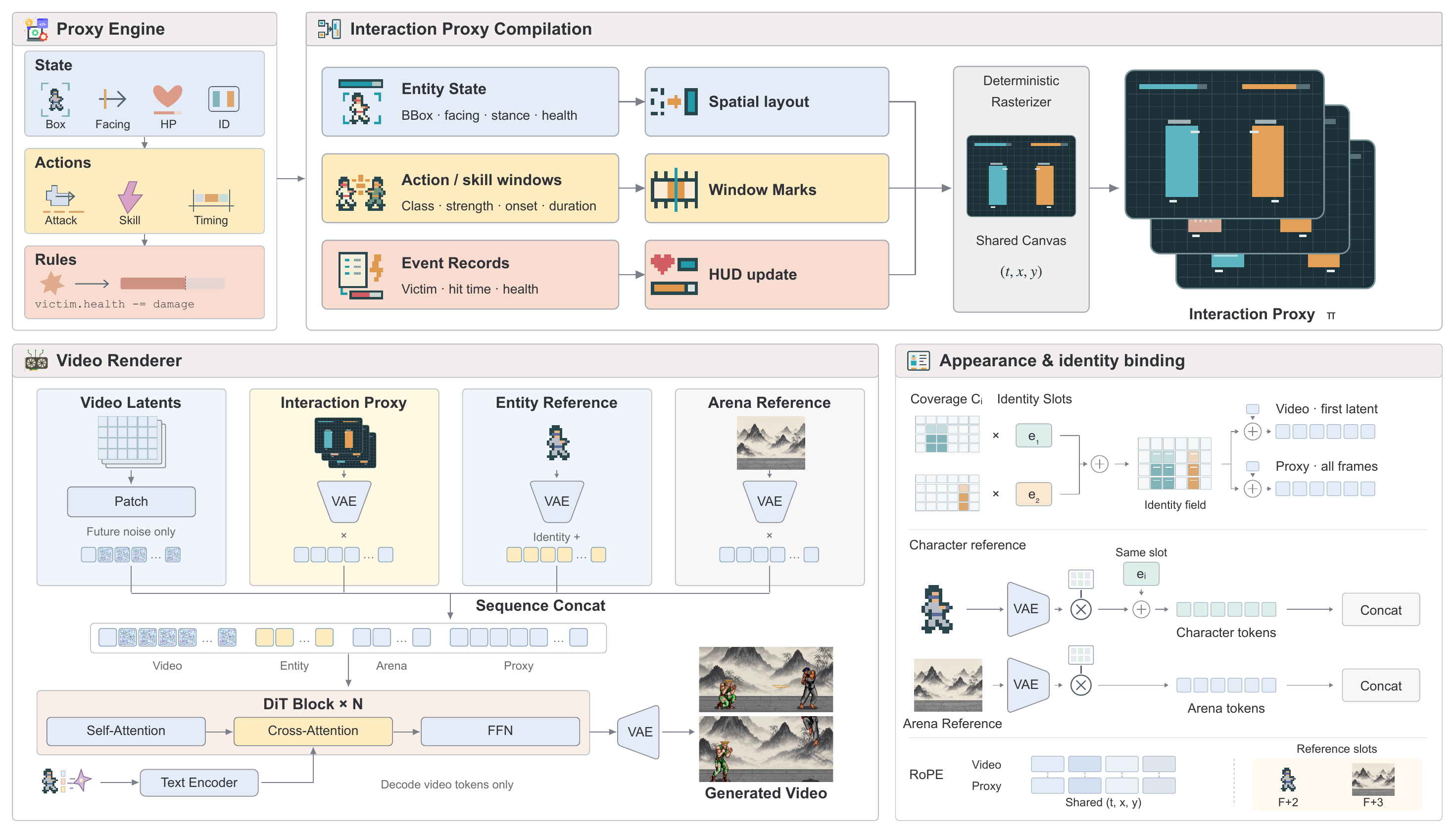}
  \caption{\textbf{Overview of FWM.} The Proxy Engine executes gameplay rules, and the resulting state trajectories, action windows, and events are compiled into a visual interaction proxy. The proxy and appearance references are encoded with a shared VAE and concatenated with video tokens. Matching spatiotemporal rotary positions align the proxy with the video, while shared identity embeddings bind character references to the corresponding regions in the first video latent and proxy sequence. The DiT processes the combined sequence, and only video tokens are decoded to generate the continuation.}
  \label{fig:overview}
\end{figure*}

We introduce Fight World Model (FWM), a customizable world-model framework that connects executable combat rules to video generation through a visual interaction proxy. As shown in \cref{fig:overview}, an executable engine resolves interactions and maintains state. A deterministic compiler rasterizes fighter states, action and skill windows, and transient hit-reaction cues onto a shared canvas and timeline. Reactions and health updates are encoded separately, allowing a hit to trigger a visible response without necessarily reducing health. Rule edits change the executed records and the resulting proxy, while the proxy format and renderer remain unchanged. Character and arena references specify appearance, with identity binding associating character references with the intended fighters across side switches. This design lets the same trained renderer depict interactions under different rules and visual appearances.

We instantiate and evaluate FWM in a fixed-view two-fighter combat setting. FightWorld Corpus pairs gameplay videos with interaction proxies across 52 skill effects, 187 character-appearance variants, and 15 arenas. FWM outperforms adapted game world models in action fidelity, hit reactions, persistent health accuracy, and character appearance preservation. Under matched executed records, reference binding, and training budgets, the interaction proxy also achieves higher joint fidelity than spatial proxies augmented with text or state tokens. Without retraining the renderer, FWM supports damage scaling, limited-hit immunity, lifesteal, and multi-stage skill mechanisms beyond native Street Fighter II/III. Appearance and side-switch evaluations test whether specified interactions and reference-to-entity associations are preserved under visual changes.
We make the following contributions:
\begin{itemize}
  \item We present FWM, a customizable combat world-model system that supports editable interaction rules and character or arena appearances without retraining the renderer.
  \item We design a visual interaction proxy that separately encodes action windows, hit reactions, and persistent health with consistent identity and timing. Identity binding connects appearance references to the fighters participating in these interactions.
  \item We construct FightWorld Corpus and evaluate the system through comparisons with existing methods, controlled encoding studies, rule interventions, and appearance customization.
\end{itemize}

\section{Related Work}
\label{sec:related}

\noindent \textbf{Game World Models.} Action-conditioned video world models predict future frames from user inputs~\cite{bruce2024genie,valevski2024gamengen,alonso2024diamond}. Recent systems extend this approach toward real-time streaming, longer horizons, and broader environments~\cite{decart2024oasis,yu2025gamefactory,zhang2025matrixgame,he2025matrixgame2,wang2026matrixgame3,robbyant2026lingbotworld,gao2026lingbotworld2,chen2026h3world}. Beyond navigation and player-centric interaction, multi-entity control introduces the need to associate each input with the intended character. Per-player keyboard streams can be paired with character references~\cite{hu2026wantofight}, while language conditions, persistent subject tokens, and synchronized inputs support multiple subjects or players~\cite{zhu2026incantation,pondaven2026actionparty,savva2026solaris}. Compact state representations also support NPC planning~\cite{deng2026worldmind}. ReactiveGWM v1 separates player keyboard control from NPC strategy conditioning~\cite{wang2026reactivegwm_v1}, and ReactiveGWM v2 uses identity handles to direct action and rule text to either fighter~\cite{wang2026reactivegwm_v2}. StatePlay additionally predicts health and skill meters alongside video through a dedicated state branch~\cite{lin2026stateplay}. These methods improve action specification, entity control, and state prediction, but interaction outcomes remain largely governed by learned video dynamics. We focus on rendered reactions and persistent state under externally edited rules.

\noindent \textbf{World State and Visual Proxies.} Spatial and trajectory conditioning provides explicit control over object placement and motion in video generation~\cite{zhang2023controlnet,wang2024boximator,luo2024ctrlv,wang2023motionctrl}. Recent world models separate state execution or maintenance from video rendering~\cite{chen2026cwm,huang2026pwm,zhan2026magpie,yang2026oneira,cai2026mass}. MASS maintains a learned authoritative state shared across players~\cite{cai2026mass}, while Magpie translates white-box engine frames into generated video~\cite{zhan2026magpie}. Oneira represents geometry and coarse interaction states through colored 3D boxes~\cite{yang2026oneira}. Code World Model compiles depth, semantic-ID, and instance maps from executable code~\cite{chen2026cwm}, and Programmable World Model projects 3D boxes into identity, semantic-category, and motion-direction maps~\cite{huang2026pwm}. These methods establish explicit interfaces between world state and visual generation. Our focus is on the interaction information needed when edited rules change the relationship between attempted actions, hit reactions, and persistent health.

\noindent \textbf{Reference-Guided Video Generation.} Reference-guided video generators condition diffusion models on reference images to preserve subject appearance across frames~\cite{liu2025phantom,jiang2025vace,zhong2025concatid,xu2025hunyuanportrait,hong2025audio}. Multi-subject methods use spatial masks, semantic alignment from vision-language models, or binding slots to distinguish appearances across entities~\cite{li2026msr,zhou2026aura,yan2026scail2}. Character animation and motion-transfer methods drive reference subjects through explicit pose sequences~\cite{hu2023animateanyone} or in-context spatiotemporal tokens aligned with the output video~\cite{luo2026dreamactorm2,yan2026scail2}. In our setting, appearance references must remain associated with the fighters performing each action and receiving each consequence, including through side switches. We study this association jointly with rule-controlled interaction outcomes.

\section{Method}
\label{sec:method}

FWM supports rule and appearance customization by coupling executable state updates with conditional video generation. As shown in \cref{fig:overview}, an engine executes the active rules, and a deterministic compiler converts the resulting states and events into a visual interaction proxy. Rule edits change the executed interaction and proxy, while appearance edits retain the proxy and change the references. Both use the same trained renderer.

\subsection{Rule Execution and Customization}
\label{sec:execution}

Let $S_k$ be the executable state at engine update $k$, $u_k$ the available action inputs, and $R$ the rule configuration. The engine computes
\begin{equation}
  (S_{k+1},e_{k+1})=F_R(S_k,u_k),
  \label{eq:transition}
\end{equation}
where $e_{k+1}$ contains the events produced together with the updated state $S_{k+1}$. The engine computes entity motion, resolves hit events, and applies the configured state updates. $S_k$ includes persistent entity attributes and the counters or timers needed to execute the rules. The configuration $R$ can be edited in code~\cite{chen2026cwm,huang2026pwm}.

We reserve $t$ for output video frames. Engine states and event records are aligned to this frame timeline before compilation. Let $\bar S_t$ denote the aligned state and $s_t=P(\bar S_t)$ its visible projection. From the aligned event history, we form active records $E_t$ that specify entity identities, onsets, and durations for action windows and reaction events. An event onset is paired with the state update it produces. The compiler then generates each proxy frame as $\pi_t=D(s_t,E_t)$.

A rule edit replaces $R$ with $R'$ and reruns the engine from the same initial state and submitted action sequence. Entity trajectories, hit times, and rule-dependent state are recomputed under $R'$. Multi-stage skill mechanisms are expressed through rules that schedule displacement, impacts, and status windows. The resulting records are compiled and rendered without changing the renderer parameters. For damage scaling, limited-hit immunity, and lifesteal, the appearance references and text conditions also remain fixed. Changes in executed outcomes reach the renderer only through the recomputed proxy.

An appearance edit changes the character or arena references while keeping the executed records and proxy fixed. Character references retain their target entity identities. Rule and appearance edits can also be combined by rendering recomputed records with new appearance conditions. For successive action intervals, the engine carries its state forward, while visual context supports continuity between generated segments.

\subsection{Interaction Proxy}
\label{sec:proxy}
\label{sec:representation}

The proxy separates attempted actions, transient reactions, and persistent state because editable rules can change their relationship. These signals share entity identities and frame indices, allowing their values and durations to vary without changing the conditioning format. We describe the concrete combat encoding used in our experiments below.

For this instantiation, the visible state at video frame $t$ is
\begin{equation}
  s_t=\{(i,b_t^i,f_t^i,q_t^i,a_t^i,h_t^i)\}_{i=1}^{2},
  \label{eq:state}
\end{equation}
where $i$ is a persistent fighter identity, and $b$, $f$, $q$, $a$, and $h$ denote its bounding box, facing direction, stance, coarse action label, and health. Each fighter's box and HUD health bar share an identity color. Facing and stance marks encode orientation and posture, and a separate band encodes the coarse action label, including an unknown value for unannotated actions.

\noindent\textbf{Action and Skill Windows.}
For the fighter with logged inputs, six attack classes combine punches and kicks with light, medium, and heavy strengths. Horizontal stripes encode punches and vertical stripes encode kicks, with coverage indicating strength. The marks remain active during the command window and supplement the coarse labels available for both fighters. They encode attempted actions, with hit outcomes specified separately. At inference, the engine can assign command windows to either fighter through its persistent identity.

A magenta block in front of the caster follows its position and facing direction to mark a skill's casting window. Text describes the effect's type and appearance, so the corpus's 52 visual skill effects share the same casting marker.

\noindent\textbf{Reactions and Persistent State.}
Reaction onset is marked by briefly inverting the affected fighter's box fill and health-bar color. Bar length independently encodes health. This supports a protected hit with a reaction cue but no health decrease, or lifesteal with an additional update to the attacker's health. Identity colors associate each reaction and state update with the affected fighter across side switches.

The same compiler processes video-derived annotations during training and engine-generated records at inference. Training reaction onsets are inferred from observed health decreases, while the engine specifies reaction events and health under the edited rules. The shared format therefore permits rule-dependent cue combinations beyond those implied by the training annotations. Rule-intervention experiments test whether the trained renderer realizes these combinations. Broad command stripes and four-frame inversions are designed to retain cues through spatial and temporal compression by the video variational autoencoder (VAE)~\cite{kingma2014vae,kong2024hunyuanvideo}. Detailed motion, textures, and effect appearance are supplied by the renderer.

\subsection{Video Rendering and Identity Binding}
\label{sec:rendering}

We condition a pretrained video diffusion transformer on the proxy and appearance inputs. Video, reference, and proxy inputs share a VAE and patch embedding. Their tokens are concatenated in video--reference--proxy order, with learned channel-wise gates on the reference and proxy streams. The initial video context, proxy, and references remain clean, while future video latents are denoised. Text descriptions are encoded by the text encoder and supplied through cross-attention. Only video tokens are decoded.

Let $x^{\mathrm{init}}=x_0$ be the initial context frame and $T$ the number of future video frames. The renderer samples
\begin{equation}
  \hat x_{1:T}\sim p_\theta(\,\cdot\mid x^{\mathrm{init}},\pi_{0:T},r,c),
  \label{eq:renderer}
\end{equation}
where $r$ contains the available character or arena references and the target identity associated with each character reference. The text condition $c$ describes appearance and skill-effect type. The proxy supplies executed interaction states and event schedules.

Proxy tokens share spatiotemporal rotary positions with the corresponding video tokens~\cite{su2024roformer}. Reference tokens use positions outside the video grid, distinguishing appearance conditions from video-aligned interaction cues.

As illustrated in \cref{fig:overview}, shared identity embeddings bind character references to persistent entity identities. For each training sample, we assign the fighters distinct slots sampled without replacement from a learned codebook. Each fighter's slot embedding is added to its corresponding tokens in the first video latent and throughout the proxy sequence, weighted by soft bounding-box coverage. The same embedding is added to all tokens of that fighter's reference. Random slot assignment discourages a fixed mapping between codebook entries and appearances. At inference, the target identity attached to a character reference selects its slot embedding. Arena references do not use an entity slot.
A training sample for character appearance editing contains a source initial image, a reference crop, and an edited target video. The source image establishes the scene and entity locations, while the reference specifies the target appearance in the continuation.

\subsection{Training}
\label{sec:training}

We optimize the diffusion transformer, conditioning gates, and identity codebook with flow matching~\cite{lipman2023flowmatching,liu2023rectifiedflow}. The objective in \cref{eq:training} supervises only future video latents, while the VAE and text encoder remain frozen. We apply condition dropout~\cite{ho2022classifierfree} to proxy, reference, and text inputs. For reference-conditioned samples, appearance-description dropout encourages reliance on the reference image.

\subsection{FightWorld Corpus}
\label{sec:data}

\begin{figure}[t]
\centering
\includegraphics[width=\linewidth]{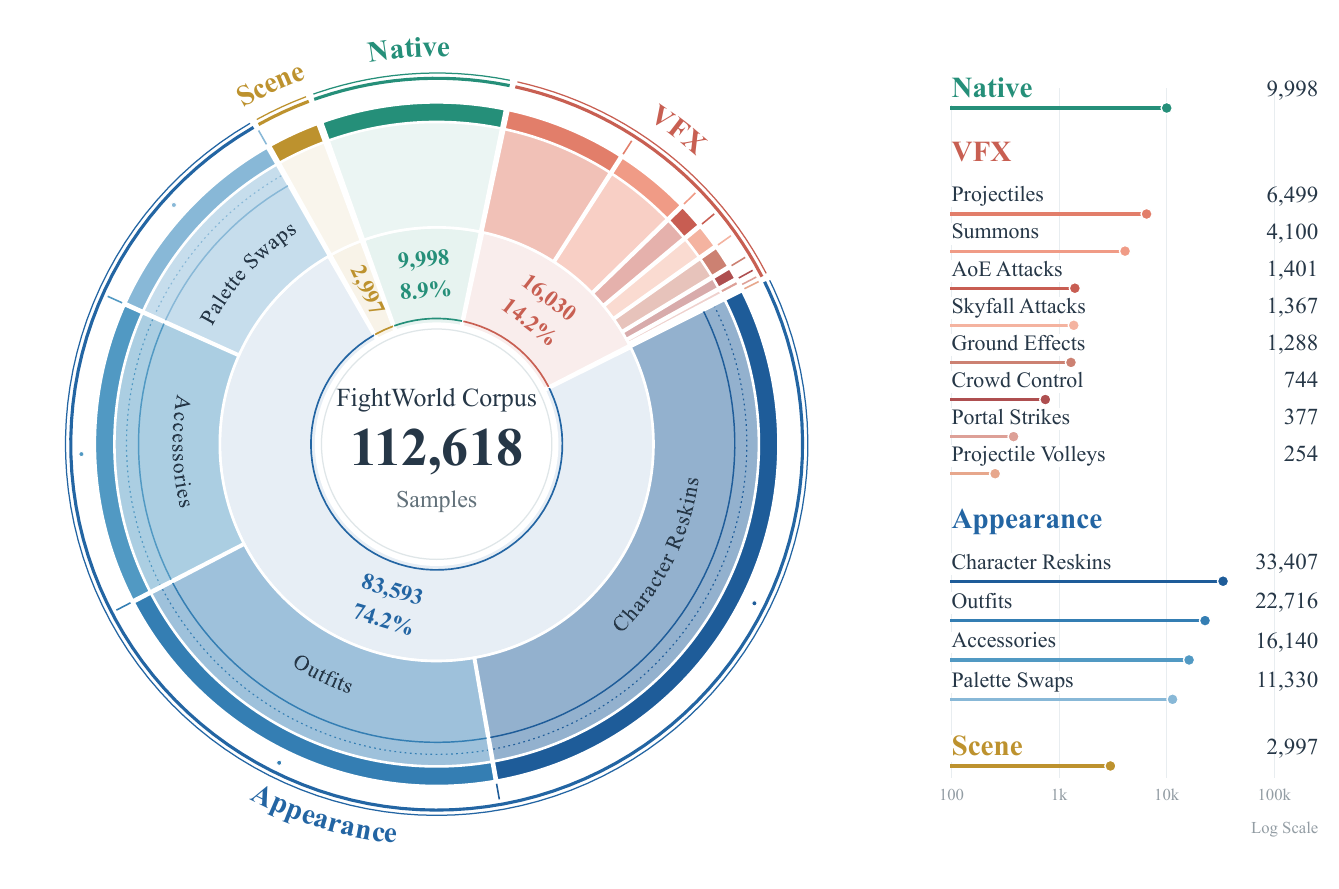}
\caption{\textbf{FightWorld Corpus.} Native gameplay is expanded with skill effects, character-appearance edits, and scene edits, yielding 112,618 training samples.}
\label{fig:corpus}
\end{figure}

We construct FightWorld Corpus by pairing Street Fighter II~\cite{capcom1991sf2} gameplay videos with interaction proxies and optional appearance references. A frozen video editor augments native recordings with edits to skill effects, character appearances, and arenas, yielding 112,618 training samples. As summarized in \cref{fig:corpus}, the corpus spans 52 skill effects, 187 character-appearance variants, and 15 arenas.

\begin{figure*}[!tp]
  \centering
  \includegraphics[width=\textwidth]{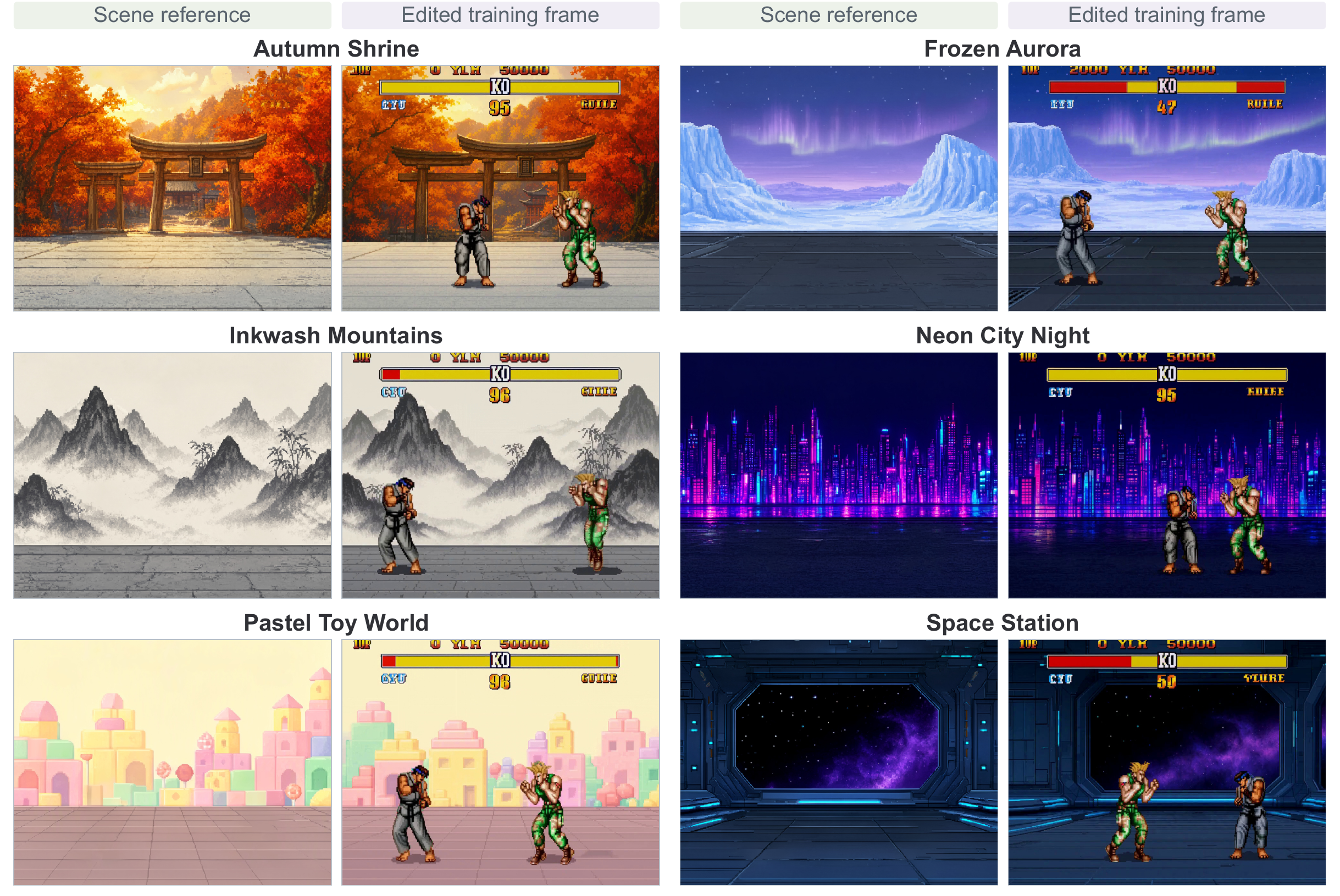}
  \caption{\textbf{Scene Editing in FightWorld Corpus.} Each scene pairs a clean reference with an edited training frame. References exclude fighters and HUD elements and may differ in background details.}
  \label{fig:scene_editing}
\end{figure*}

Character and arena edits vary visual appearance while retaining the source interaction. We reuse source state annotations only after checking consistency in fighter trajectories, event timing, and health. Each accepted video is paired with a character crop or an arena reference, as illustrated for scene edits in \cref{fig:scene_editing}.

Skill edits change the visual effect associated with a casting event and may also alter the visible interaction. We therefore extract state and event annotations from both native and skill-edited videos, retaining only clips that pass quality screening.

Native and edited videos share the same training format, with interaction proxies, text descriptions, and optional character or arena references as conditions. We use these samples to train a shared renderer across skill effects and appearances.

\section{Experiments}
\label{sec:experiments}

We evaluate appearance control, interaction fidelity, and rule customization. We compare FWM with adapted game world models and conduct controlled studies to examine the contributions of interaction encoding and identity binding.

\begin{figure*}[!tp]
  \centering
  \includegraphics[width=\referencegalleryfigwidth]{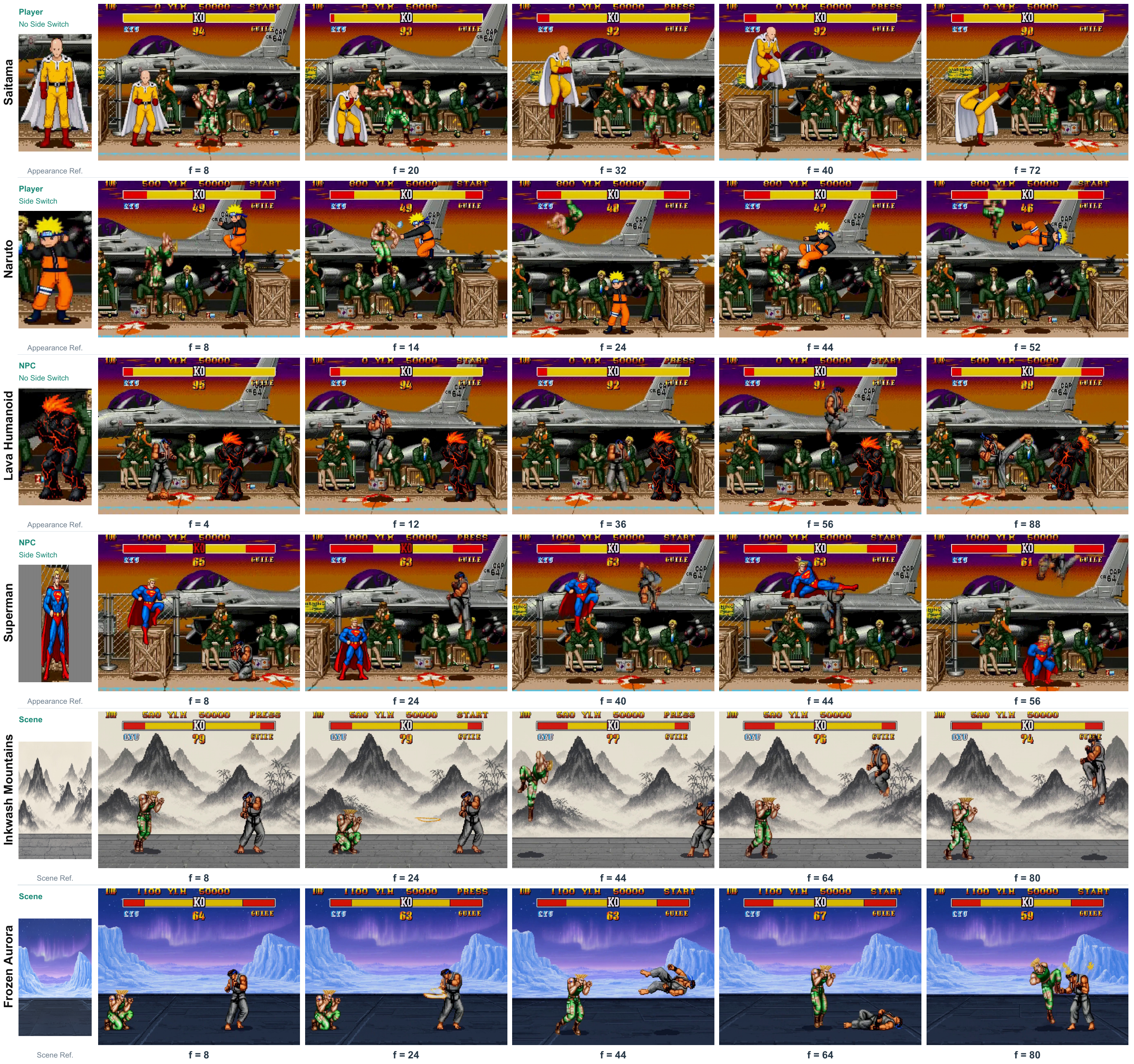}
  \caption{\textbf{Character and Scene Reference Conditioning.} Interaction rollouts across four character appearances (under default and side-switched stances) and two arena references.}
  \label{fig:appearance_ref}
\end{figure*}

\begin{figure*}[!tp]
  \centering
  \includegraphics[width=\resultfigwidth]{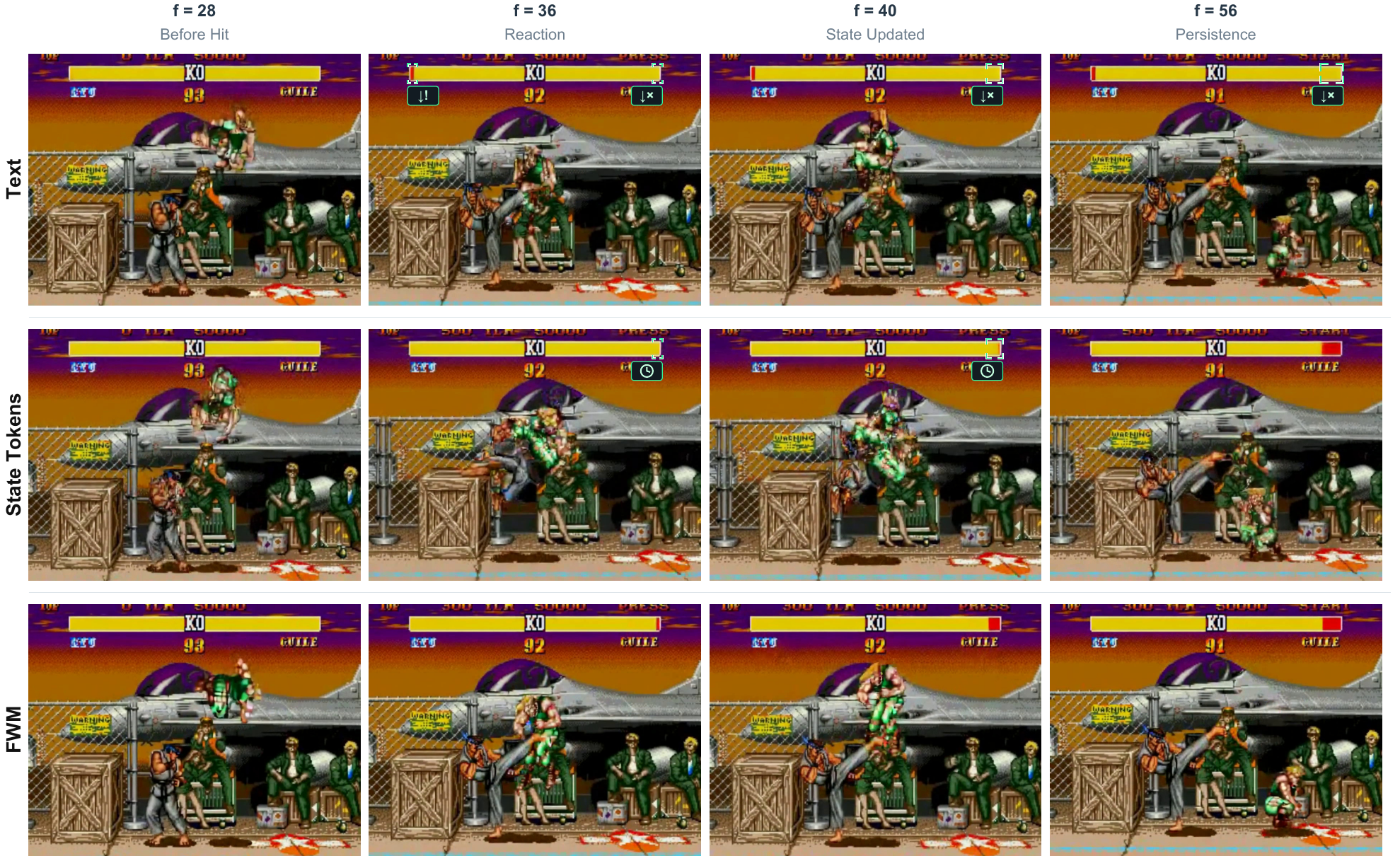}
  \caption{\textbf{Complete-Information Encodings on the Same Interaction.} Held-out rollout under matched executed records (\cref{tab:main}). Green brackets mark HUD state errors: \icondrop{} unexpected health loss, \iconmiss{} missing health loss, \icondelay{} delayed update, \iconpart{} incomplete update, and \iconrise{} unexpected recovery.}
  \label{fig:state_rep}
\end{figure*}

\begin{figure*}[!tp]
  \centering
  \includegraphics[width=\resultfigwidth]{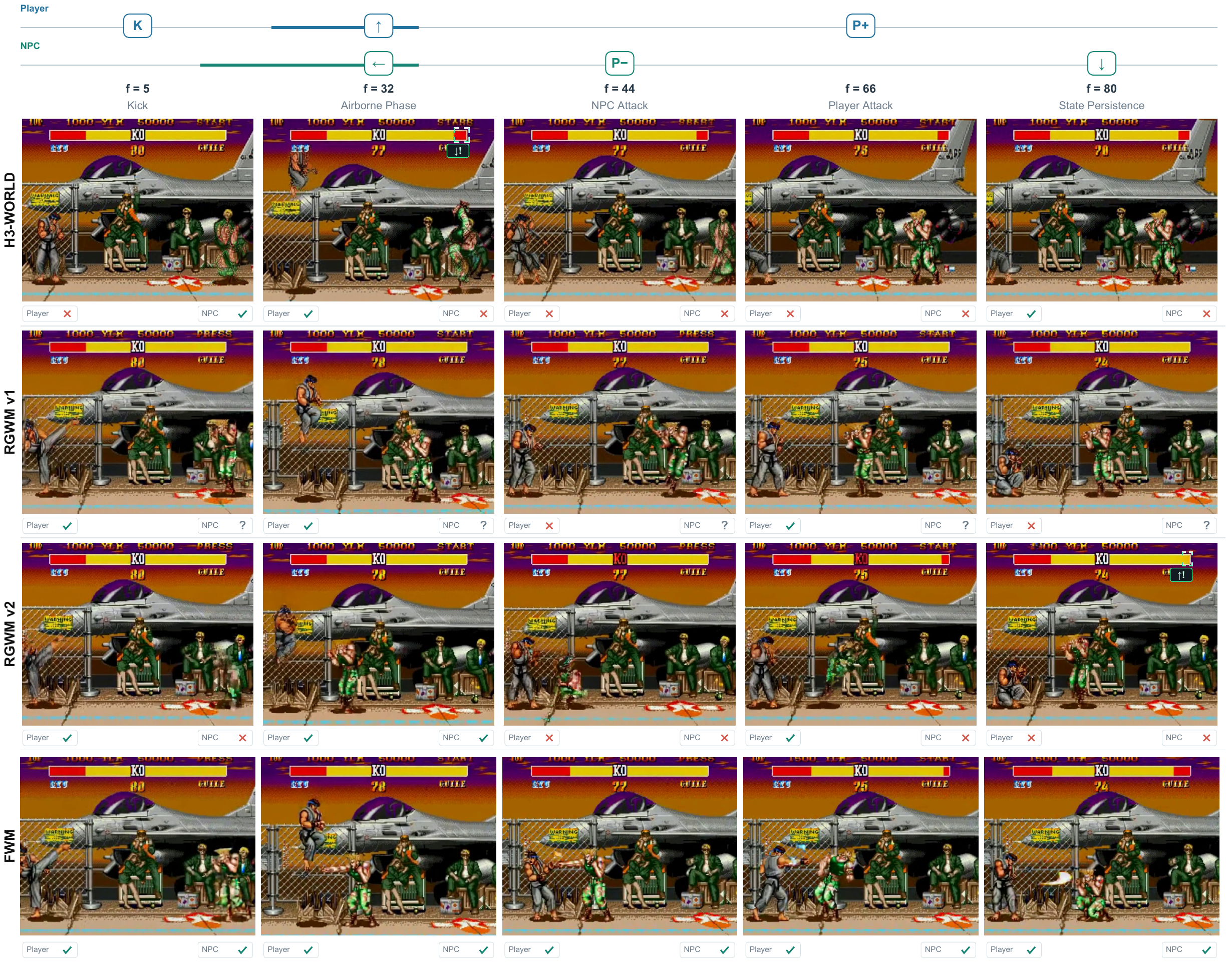}
  \caption{\textbf{Dual-Character Actions and Responses.} Top rails show visible action summaries for both fighters. Per-frame badges indicate realized (\verdictok{}) or unrealized (\verdictbad{}) fighter actions and responses; \verdictunk{} denotes strategy-level NPC text in ReactiveGWM v1; HUD state icons follow \cref{fig:state_rep}.}
  \label{fig:dual_actions}
\end{figure*}

\begin{figure*}[!tp]
  \centering
  \includegraphics[width=\resultfigwidth]{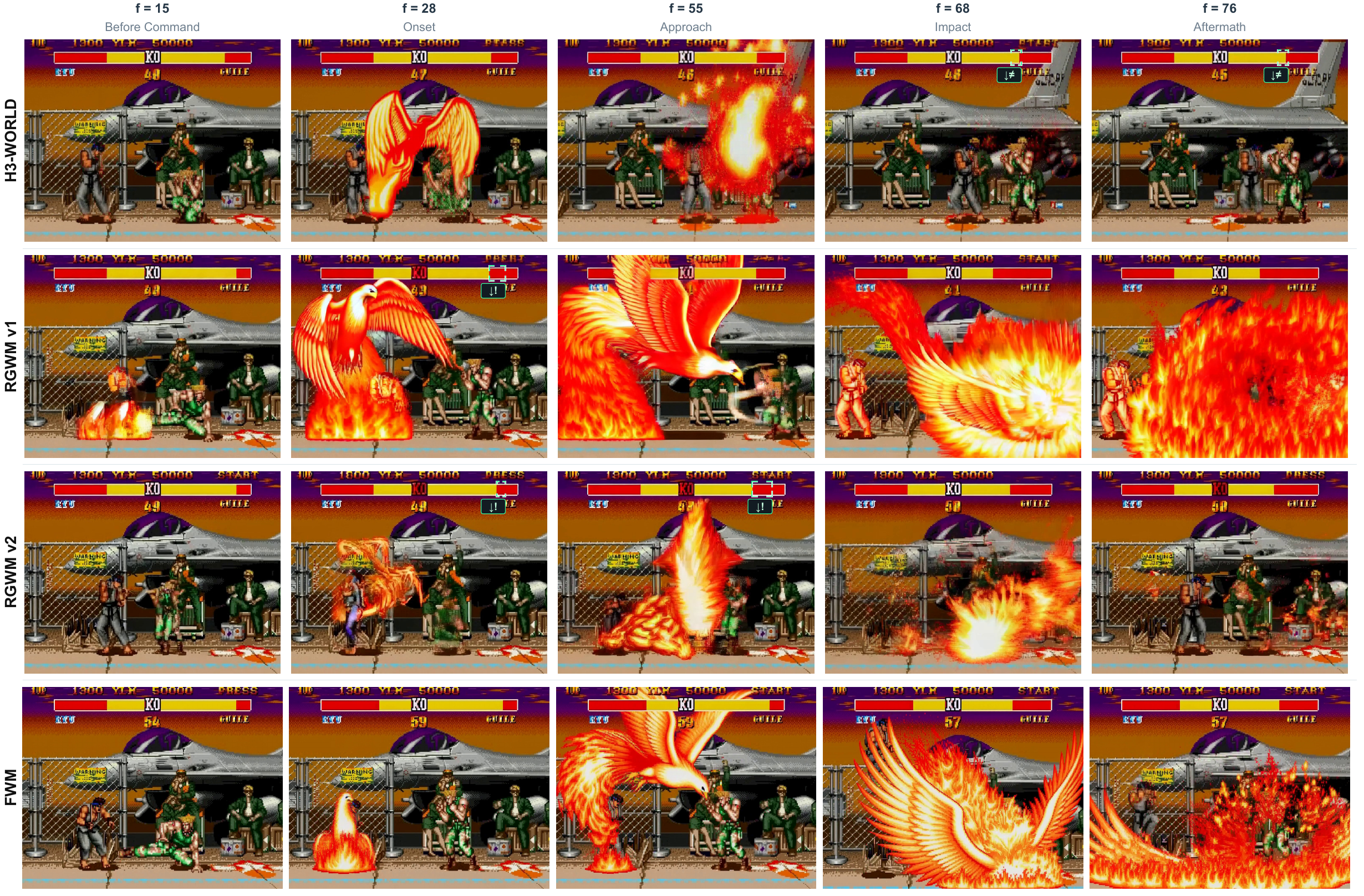}
  \caption{\textbf{Skill Timing and Event Consequences.} Comparison on a projectile skill commanded at $f=20$; HUD state icons follow \cref{fig:state_rep}.}
  \label{fig:skill_timing}
\end{figure*}

\input{tables/04_reference_binding.tex}
\input{tables/01_control_representations.tex}
\input{tables/02_existing_methods.tex}
\input{tables/05_proxy_ablations.tex}
\input{tables/03_rule_interventions.tex}

\subsection{Evaluation Setup}

\label{sec:setup}

\label{sec:metrics}

\noindent\textbf{Implementation.}
We evaluate FWM on two-fighter interactions in the fixed 2D view of Street Fighter II~\cite{capcom1991sf2}. Our renderer uses Wan2.2-TI2V-5B~\cite{wan2025wan} as its backbone and is trained on FightWorld Corpus.

\noindent\textbf{Evaluation Protocol.}
For recorded interactions, held-out video annotations specify the target interactions. For rule interventions, we fix the initial state and submitted actions, then rerun the engine under each rule configuration to compute trajectories, hit events, and state updates. FWM's reference images and text condition $c$ remain unchanged for damage scaling, limited-hit immunity, and lifesteal. Appearance evaluations change the references while retaining the executed interaction. Each controlled encoding comparison supplies identical interaction records to all variants.

\noindent\textbf{Action and Reaction Metrics.}
Move Accuracy (MoveAcc) measures agreement between fighter motion and directional commands. Attack Accuracy (AttackAcc) is the fraction of detected punch or kick events supported by the corresponding input commands within their temporal windows. Skill Accuracy (SkillAcc) requires the correct skill class and an onset within the command-relative evaluation window. Reaction measures the fraction of expected hit responses visibly performed by the designated fighter. Qwen judges these responses from body motion in nearby frames, with the HUD excluded.

\noindent\textbf{Health and Persistence.}
Health MAE measures the mean absolute error between generated and target HUD health, normalized by maximum health and reported in percentage points. The joint health check requires correct updates for both fighters and persistence of the resulting health values between events.

\noindent\textbf{Joint Interaction Fidelity.}
Joint interaction fidelity ($J$) treats an attempt as successful only if all applicable action, entity, timing, reaction, and health checks pass. We average attempt-level results within each episode and give episodes equal weight. For rule interventions, $J_{\mathrm{rule}}$ requires the applicable reaction and health checks to pass together.

\noindent\textbf{Appearance and Entity Assignment.}
Adherence uses a VLM to judge whether the requested appearance is applied to the designated character or arena. The judge receives the target reference and a source-frame identity guide. DINO measures cosine similarity between reference and generated features for characters and arenas separately. Both metrics exclude the fixed initial context.

\noindent\textbf{Baselines.}
We compare FWM with H3-WORLD~\cite{chen2026h3world}, ReactiveGWM v1~\cite{wang2026reactivegwm_v1}, and ReactiveGWM v2~\cite{wang2026reactivegwm_v2}. We adapt all three methods to FightWorld Corpus and its 52 skill effects. For skill control, we add a dedicated trigger to v1 and v2 and specify activation in H3-WORLD's interval action text. The +Ref variants use our reference-conditioning branch for appearance editing.

\subsection{Character and Scene Customization}
\label{sec:binding_results}

As illustrated in \cref{fig:appearance_ref}, FWM renders the same prescribed interaction with different character and arena references. We retain the executed records and proxy while changing appearance, testing whether the specified actions and consequences are preserved.

\noindent\textbf{Identity Association across Side Switches.}
We further test whether character references remain associated with their intended fighters across side switches. This requires preserving identity independently of the fighters' screen positions. The variants in \cref{tab:binding_control} share reference inputs and training budgets, differing only in whether identity binding is enabled. Binding improves character adherence and DINO similarity, with an adherence gain of approximately 45 percentage points on side-switch clips.

\subsection{Control Representation Comparisons}

\label{sec:control_comparison}

\noindent\textbf{From Commands to Executed Outcomes.}
Adding rule text to action commands improves joint fidelity, but the generated interactions remain inconsistent. Spatial Proxy Only accurately controls movement while showing low joint fidelity in \cref{tab:main}. These conditions leave the renderer to infer hit events and health updates. The results motivate supplying executed outcomes alongside attempted actions.

\noindent\textbf{Holding Executed Information Fixed.}
To isolate the effect of encoding, we compare Spatial Proxy + Text, Spatial Proxy + State Tokens, and FWM under identical executed interaction records. The records specify fighter identities, action classes, affected fighters, event onsets and durations, and frame-wise health trajectories. All variants share geometry, initial context, reference binding, backbone initialization, and training budgets.

\noindent\textbf{Effect of Interaction Encoding.}
With identical executed information, FWM improves joint fidelity by approximately 22 percentage points over Spatial Proxy + Text, the stronger symbolic baseline in \cref{tab:main}. State Tokens have lower average health error than Text but poorer joint fidelity, showing that average health accuracy does not fully capture interaction correctness.

Text misassigns damage to the attacker or misses the recipient's health update in \cref{fig:state_rep}. State Tokens track health more closely on average but lag at impact onset and show local limb blur during contact. Both baselines supply events and health as symbolic records that must be associated with body regions, HUD bars, and video frames. FWM places these cues in the corresponding proxy regions and frames.

\subsection{Comparison with Existing Methods}

\label{sec:method_comparison}

\begin{figure*}[!t]
  \centering
  \includegraphics[width=\tallresultfigwidth]{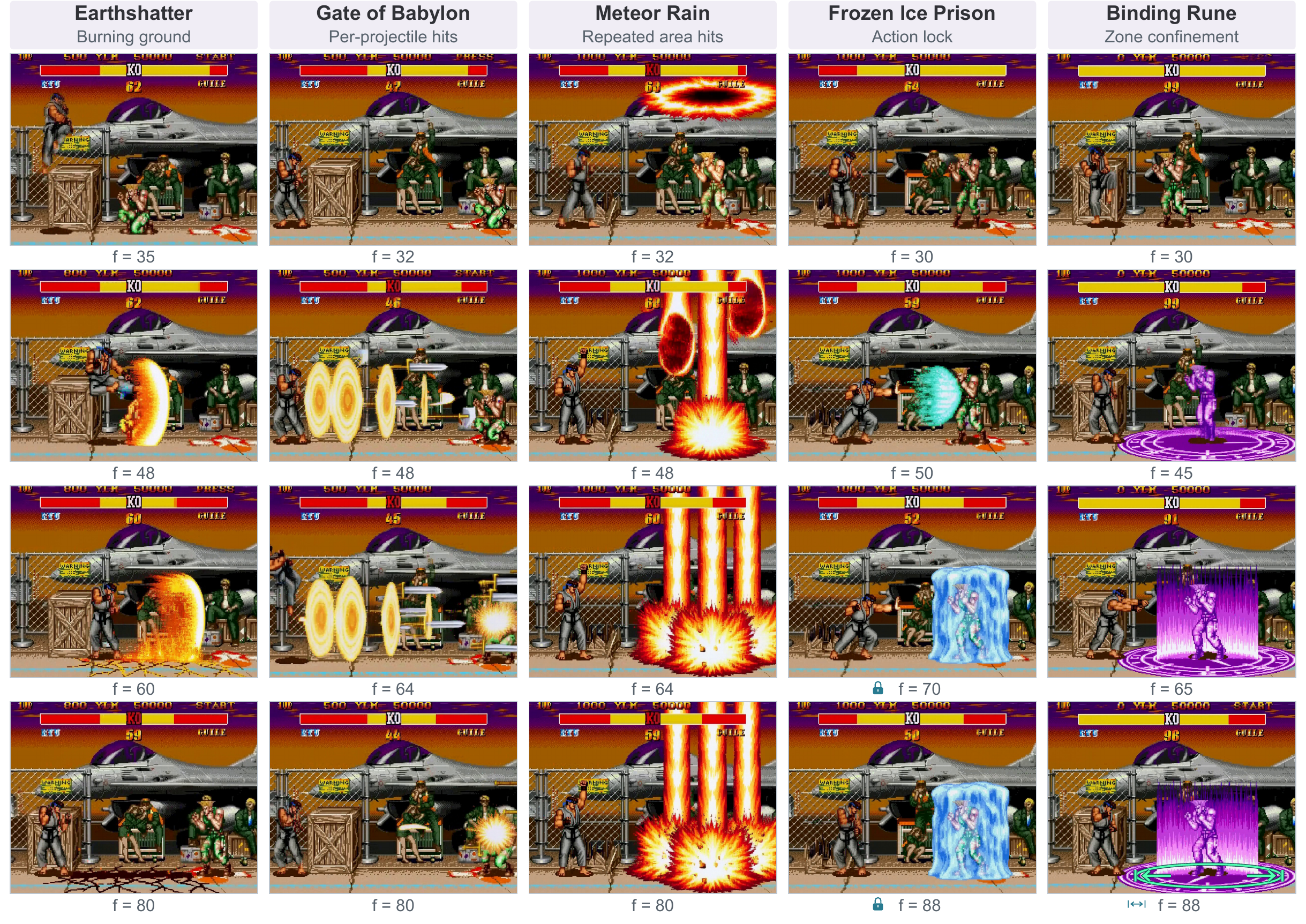}
  \caption{\textbf{Skill-Level Rule Mechanisms Beyond Native Combat.} Five combat mechanisms absent from native \emph{Street Fighter II/III}, spanning lingering ground damage, per-projectile hits, staggered area hits, action locks, and spatial confinement (boundary locked).}
  \label{fig:skill_mechanisms}
\end{figure*}

\noindent\textbf{Action Control and State Consistency.}
ReactiveGWM v1~\cite{wang2026reactivegwm_v1} achieves higher player movement and attack precision than H3-WORLD~\cite{chen2026h3world} in \cref{tab:method_comparison}. It uses keyboard inputs for the player and coarser strategy text for the NPC. ReactiveGWM v2~\cite{wang2026reactivegwm_v2} directs action instructions to either fighter, but still infers hit outcomes and health updates within video generation. The baselines miss actions and reactions in \cref{fig:dual_actions}, and v2 shows spurious health recovery.

\noindent\textbf{Multi-Phase Skills.}
The baselines struggle to synchronize skill effects with their consequences. In \cref{fig:skill_timing}, ReactiveGWM v1 and v2 deduct health before projectile impact, while H3-WORLD shows insufficient health loss after contact. ReactiveGWM v2 also renders a malformed projectile effect.

\noindent\textbf{Appearance and Interaction Fidelity.}
Adding appearance references improves v2's movement and skill accuracy by approximately 22 and 18 percentage points, respectively. The +Ref baselines achieve high character adherence in \cref{tab:method_comparison}, yet their hit reactions and health updates remain inconsistent. Garment blur and identity drift during rapid contact and side switches further indicate that overall adherence does not guarantee stable visual identity throughout an interaction.

\noindent\textbf{Executed Outcomes as Visual Conditions.}
FWM improves action accuracy, hit reactions, health consistency, and character adherence over the adapted baselines in \cref{tab:method_comparison}. Its Reaction score exceeds the best +Ref result by approximately 43 percentage points. The rollouts in \cref{fig:dual_actions,fig:skill_timing} follow the prescribed action and reaction timing while preserving the resulting health across subsequent frames.

\subsection{Rule Interventions}

\label{sec:interventions}

\noindent\textbf{Rule and Skill Customization.}
We test whether generated interactions follow edited rules and newly composed skill mechanisms without retraining the renderer. For attack-level interventions, the initial state, submitted actions, appearance references, and FWM text condition $c$ remain fixed. The engine recomputes motion, hit timing, and state updates, and the changed outcomes reach FWM only through the proxy. FWM maintains high joint fidelity across the revised rules in \cref{tab:rules}, while symbolic encodings degrade particularly under immunity and lifesteal.

\noindent\textbf{Damage Magnitude.}
Damage scaling changes the health loss associated with a valid hit. FWM tracks the resulting health changes under both reduced and amplified damage, with low health error and high joint fidelity.

\noindent\textbf{Limited-Hit Immunity.}
Limited-hit immunity blocks health loss for a specified number of hits while retaining reaction events. FWM renders the corresponding body responses with unchanged health during protection, then shows the prescribed health loss once protection is exhausted.

\noindent\textbf{Lifesteal.}
Under lifesteal, a hit reduces the recipient's health and restores the attacker's health. FWM renders these coupled updates for the correct fighters and preserves the resulting health values after the hit.

\noindent\textbf{Multi-Stage Skill Mechanisms.}
We extend the rule interventions to skills that combine displacement, repeated impacts, and state constraints. In \cref{fig:skill_mechanisms}, Earthshatter couples knockback with staged health loss and lingering ground damage. Other mechanisms cover per-projectile hits, staggered area impacts, action locks, and spatial confinement. FWM renders these combinations through the shared proxy without additional skill-specific conditioning channels or renderer retraining.

\subsection{Proxy Ablations}

\label{sec:ablations}

\noindent\textbf{Attempted Actions and Their Consequences.}
We retrain each variant with one cue family removed, keeping the training setting unchanged. In \cref{tab:ablation}, removing command marks lowers attack precision while leaving Reaction largely unchanged. Removing the skill trigger reduces skill accuracy by approximately 20 percentage points. These results support encoding attempted actions separately from their consequences.

\noindent\textbf{Events and Persistent State.}
Removing damage-event cues lowers the Reaction score while health error remains low. Removing health bars largely preserves reactions but degrades health accuracy and joint fidelity. These complementary effects support separate conditioning for transient reactions and persistent state. More details, comparisons and visualizations are provided in the appendix.

\section{Conclusion}
\label{sec:conclusion}

We presented Fight World Model (FWM), which connects executable gameplay rules to video generation through a visual interaction proxy. Experiments in a combat setting demonstrate improved interaction and appearance fidelity over adapted game world models. FWM supports edited rules and multi-stage skill mechanisms without retraining the renderer. Character and arena customization preserves the prescribed interactions, with identity binding maintaining character associations across side switches.
{\small
\bibliographystyle{ieeenat_fullname}
\bibliography{main}
}
\appendix
\section{Implementation Details}
\label{app:implementation}
\label{sec:implementation}

\subsection{Engine and Timeline}
\label{app:engine_timeline}
The combat engine exposes damage magnitude, the number of protected hits, and the lifesteal rate as editable parameters. Its state retains protection counts and active skill windows across action intervals.

Engine updates indexed by $k$ are mapped to output video frames $t$ before proxy compilation. State samples and event windows use the same frame mapping, so an event onset is paired with its resulting state update. Frame index zero denotes the initial context, and $\hat x_{1:T}$ denotes the continuation. The engine carries executable state between action intervals, while visual context conditions the next generated segment.

\subsection{Proxy Encoding Parameters}
\label{app:proxy_encoding}
\begin{figure*}[t]
\centering
\includegraphics[width=\textwidth]{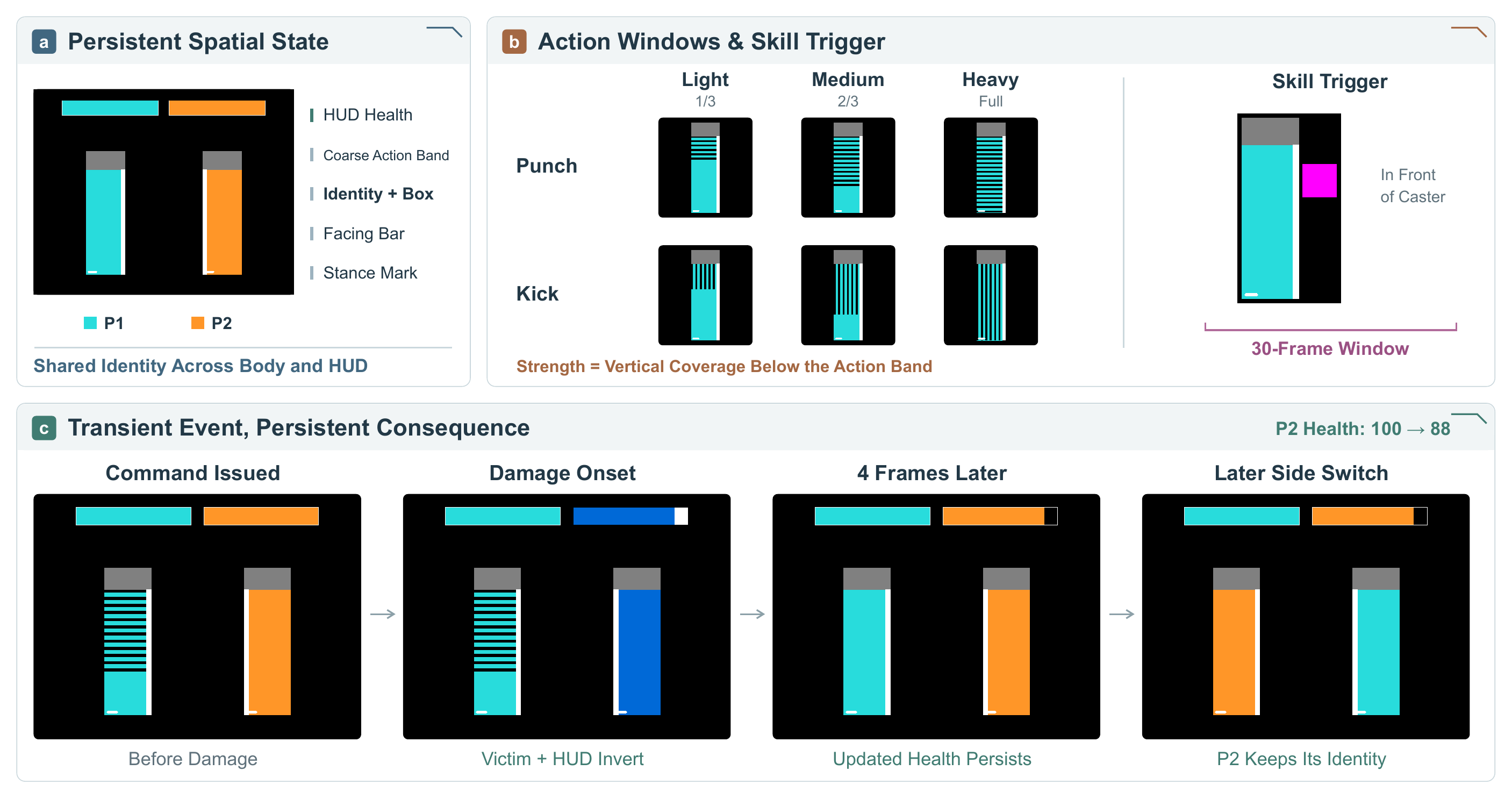}
\caption{\textbf{Interaction-Proxy Vocabulary.} Transient cues and persistent state share fighter identities.}
\label{fig:proxy}
\end{figure*}
For light, medium, and heavy attacks, stripes cover the top one third, top two thirds, or full box interior below the coarse action band. They remain active over the command window. The skill-trigger block follows the caster for 30 output video frames, and reaction color inversions last four frames. Both durations are defined at the output video rate and resampled together with the state and event records.

\subsection{Renderer and Binding Parameters}
\label{app:rendering_parameters}
We initialize the renderer from the Wan2.2-TI2V-5B checkpoint~\cite{wan2025wan}. 
For $F$ video latent frames, character references use temporal coordinate $\tau_{\mathrm{ref}}=F+2$, and arena references use $\tau_{\mathrm{ref}}=F+3$. Their spatial coordinates lie outside the video grid. Video and proxy inputs use matched temporal indices and the same resize and center-crop transform.

The identity codebook contains six learned slots. Each embedding is normalized to unit root-mean-square magnitude and scaled by 0.25 at all three injection sites. For the first video latent and proxy sequence, soft bounding-box coverage weights the corresponding entity's embedding, with weights normalized where boxes overlap. The target entity's embedding is added to all character-reference tokens without a spatial box mask. Persistent entity IDs in the proxy identify the reference targets. Arena references receive no identity embedding.

\subsection{Training Objective and Configuration}
Let $z$ denote clean video latents and $\epsilon\sim\mathcal N(0,I)$. We construct noisy latents with flow matching~\cite{lipman2023flowmatching,liu2023rectifiedflow}:
\begin{equation}
  z_\tau=(1-\tau)z+\tau\epsilon,\qquad
  \tau=\frac{5v}{1+4v},\quad v\sim\mathcal U(0,1).
\end{equation}
Here $\tau$ is flow time, distinct from engine index $k$ and video-frame index $t$. The first latent is replaced with the clean source-image encoding at noise level zero. Future video tokens receive noise level $\tau$. With conditions $\mathcal C=(x^{\mathrm{init}},\pi,r,c)$, the loss is
\begin{equation}
\mathcal L=\mathbb E\!\left[
\frac{1}{|\Omega_+|}\sum_{j\in\Omega_+}
\left\|v_\theta(z_\tau,\tau;\mathcal C)_j-(\epsilon-z)_j\right\|_2^2
\right],
\label{eq:training}
\end{equation}
where $\Omega_+$ contains future video latent elements. Initial-context, proxy, and reference tokens are excluded from the loss. At inference, future video latents start from noise.

We use AdamW with a constant learning rate of $8\times10^{-5}$, weight decay of 0.01, and no learning-rate warmup. Training uses bfloat16 autocast with fp32 master parameters and an effective batch size of 128 on 128 H200 GPUs. The budget is 35 epochs, corresponding to 30,765 optimizer steps.

Each clip contains 89 frames at $928\times704$ resolution. Dropout probabilities for proxy, reference, text, and appearance descriptions are 0.1, 0.1, 0.1, and 0.2, respectively. The controlled encoding variants in \cref{tab:main} share initialization, training data, and optimization budget. Only the representation and its required input layers differ.

\section{Data Construction Details}
\label{app:data}

\subsection{Video Editing and Quality Control}
A frozen MiniMax-H3 Ref2VA editor produces appearance and skill-effect variants using edit-specific prompts. Screening checks the requested content, temporal coherence, and interaction consistency.

Appearance-only variants are checked against the source video for fighter trajectories, event timing, and health traces. Only edits that pass these checks reuse the source state annotations. Skill-effect edits are annotated from the edited videos.

\subsection{State and Event Annotation}
Gemma-4-31B-it localizes fighters at regularly spaced keyframes to initialize SAM~2 video propagation~\cite{ravi2024sam2}. Semantic descriptions help distinguish fighters from similar background characters. At each keyframe, the tracker checks the two identity assignments against appearance and trajectory continuity, retaining or swapping them to correct drift and bridge short gaps. Bounding boxes specify spatial state. Facing direction follows relative fighter positions and retains its previous value while the boxes overlap.

Health traces are estimated from HUD bar lengths and cleaned within each round. Nearby decreases are merged into damage events, each recording the affected fighter and event time. Gemma also supplies coarse action and stance labels at selected frames, including frames near damage events. Unannotated action frames use the unknown label. Logged inputs provide fine attack commands and scheduled skill triggers for the controlled fighter. The compiler converts these annotations into interaction proxies.

\section{Evaluation Protocols}
\label{app:evaluation}

\subsection{Controlled Interactions}
Recorded-interaction tests use held-out video annotations, with source episodes separated between training and testing. Rule interventions rerun the engine from the same initial state and submitted actions under each configuration, recomputing motion trajectories, hit events, and state outcomes. These records define the target interactions for the compared encodings. In FWM's damage scaling, limited-hit immunity, and lifesteal tests, reference images and text condition $c$ remain fixed. Changed outcomes reach FWM only through the proxy.

Appearance interventions retain the executed records and proxy while changing the references. Evaluation cases include no-hit attempts, repeated attacks, overlap, and side switches.

\noindent\textbf{Complete-Information Encodings.}
The encodings use matched event-free intervals, sampled values, and temporal indices, with shared resolution and sampling settings. Each uses the input layers required by its representation, and text inputs are checked for truncation.

\subsection{Action Recognition and Timing}
\noindent\textbf{Movement.}
MoveAcc compares tracked fighter motion with directional commands over valid command windows and averages across the four movement keys. Jump positives require an actual airborne jump. Grounded hurt recovery is excluded from jump positives.

\noindent\textbf{Attack and Skill Recognition.}
We train ResNet-18-based temporal convolutional networks (TCNs) for attack and skill recognition. Attack detections are grouped into temporally contiguous events. AttackAcc measures the fraction of detected punch or kick events supported by the corresponding input command within its temporal window.

The skill classifier uses a 256-channel temporal trunk with dilation rates 1, 2, 4, and 8. It predicts onset and class, with a no-skill outcome in the onset prediction. Class logits are pooled over the predicted effect interval. SkillAcc requires the correct class and onset within the command-relative window. The same timing rule applies across methods and accommodates skill startup, release, and recovery.

An attempted attack may hit or miss, so AttackAcc and Reaction assess command-grounded action execution and received-hit consequences, respectively.

\subsection{Reaction and Health}
\noindent\textbf{Visible Reactions.}
Reaction measures the fraction of expected hit responses visibly performed by the designated fighter within the evaluation window. Qwen judges body motion in surrounding gameplay frames, using a source-frame identity guide and excluding the HUD.

Brief reactions can resemble ordinary motion across appearances and effects, so we judge them using local temporal context around expected events. Computing reaction precision would additionally require locating every visible received-hit onset throughout the generated video, a dense detection task that is difficult for current VLMs.

\noindent\textbf{Health and Persistence.}
Let $\mathcal V$ contain readable fighter-frame observations and $H$ denote maximum health. Health MAE is the normalized mean absolute error, expressed in percentage points:
\begin{equation}
  \mathrm{MAE}_h=\frac{100}{|\mathcal V|H}
  \sum_{(i,t)\in\mathcal V}|\hat h_t^i-h_t^i|.
  \label{eq:health}
\end{equation}
Observed and target health values are obtained from HUD bar lengths. The joint health check uses a one-pixel tolerance and requires correct updates for both fighters, with the resulting values preserved between events. This includes unchanged health under immunity and the attacker's restored health under lifesteal.

\subsection{Joint Interaction Fidelity}
An attempt succeeds only when all applicable action, entity, timing, reaction, and health checks pass. The score $J$ averages attempts within each episode and weights episodes equally:
\begin{equation}
  q_{\ell e}=\prod_{k\in\mathcal K_{\ell e}}c_{\ell e,k},\qquad
  J=\frac{1}{L}\sum_{\ell=1}^{L}\frac{1}{N_\ell}
  \sum_{e=1}^{N_\ell}q_{\ell e}.
  \label{eq:joint}
\end{equation}
Here $L$ counts episodes, $N_\ell$ counts attempts in episode $\ell$, and $c_{\ell e,k}$ are binary checks indexed by $\mathcal K_{\ell e}$. The applicable set follows the registered interaction. Its checks cover command-grounded actions and timing, the identities of acting and affected fighters, visible body responses, and correct health updates with persistence. For rule interventions, $J_{\mathrm{rule}}$ combines the required reaction and health checks. Both rates are expressed as percentages in the results.

\subsection{Appearance and Entity Assignment}
Adherence assesses appearance-edit success for the designated character or arena. A VLM judges sampled continuation frames using the target reference and a source-frame identity guide. Frame judgments are combined into a clip-level decision.

DINO measures reference-to-generation feature cosine similarity separately for characters and arenas. Proxy boxes localize fighter crops, while fighter and HUD regions are suppressed for arena scoring. The fixed initial context is excluded from both appearance metrics.

\section{Reference-Augmented Baselines}
\label{app:ref_baselines}

\begin{figure*}[p]
  \centering
  \includegraphics[width=\appendixfigwidth,height=0.42\textheight,keepaspectratio]{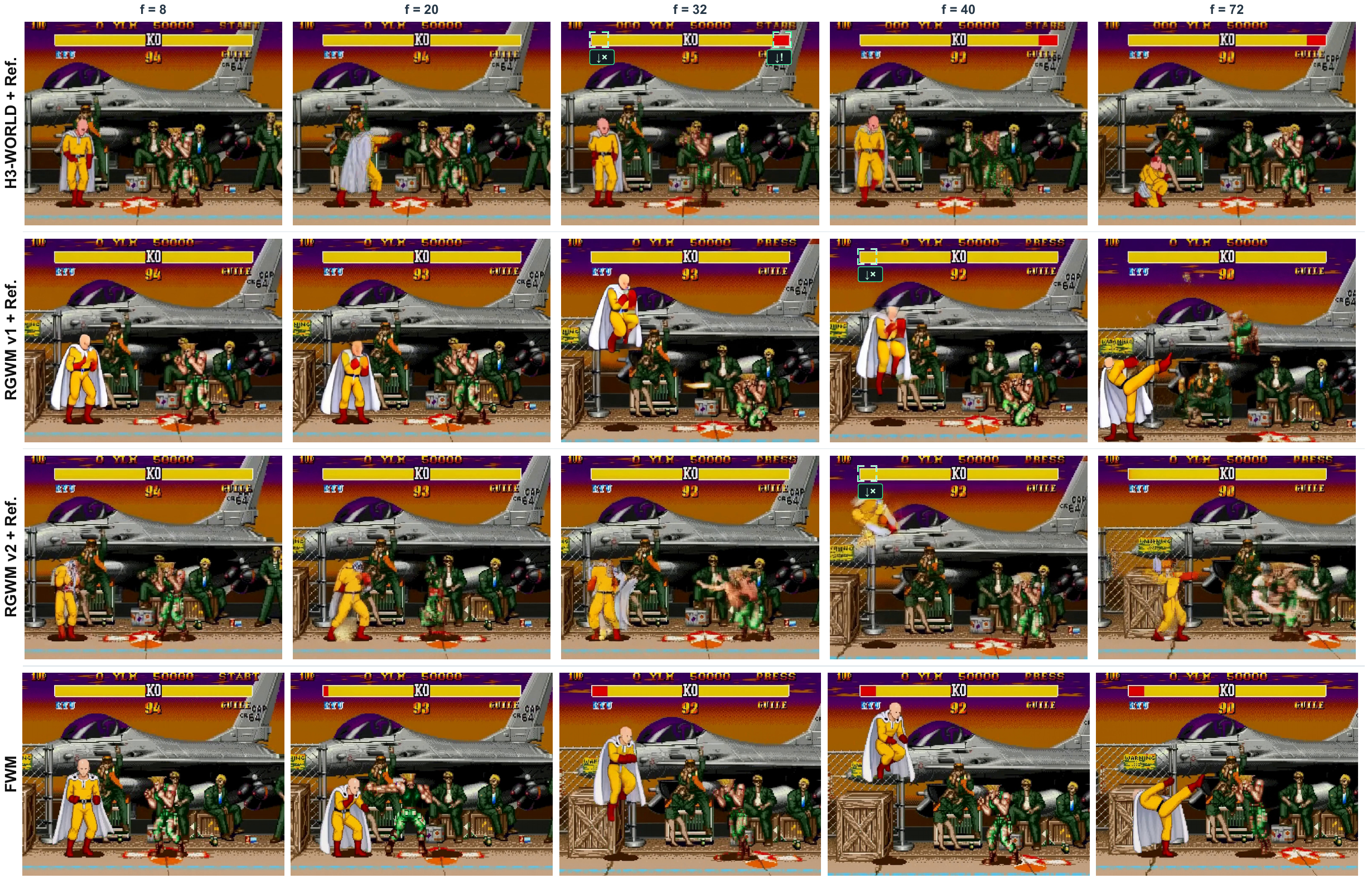}
  \caption{\textbf{Reference-Augmented Comparison (Saitama, Player, No Side Switch).} Matched frames corresponding to row 1 of \cref{fig:appearance_ref}; HUD state icons follow \cref{fig:state_rep}.}
  \label{fig:app_saitama}
  \vspace{8pt}
  \includegraphics[width=\appendixfigwidth,height=0.42\textheight,keepaspectratio]{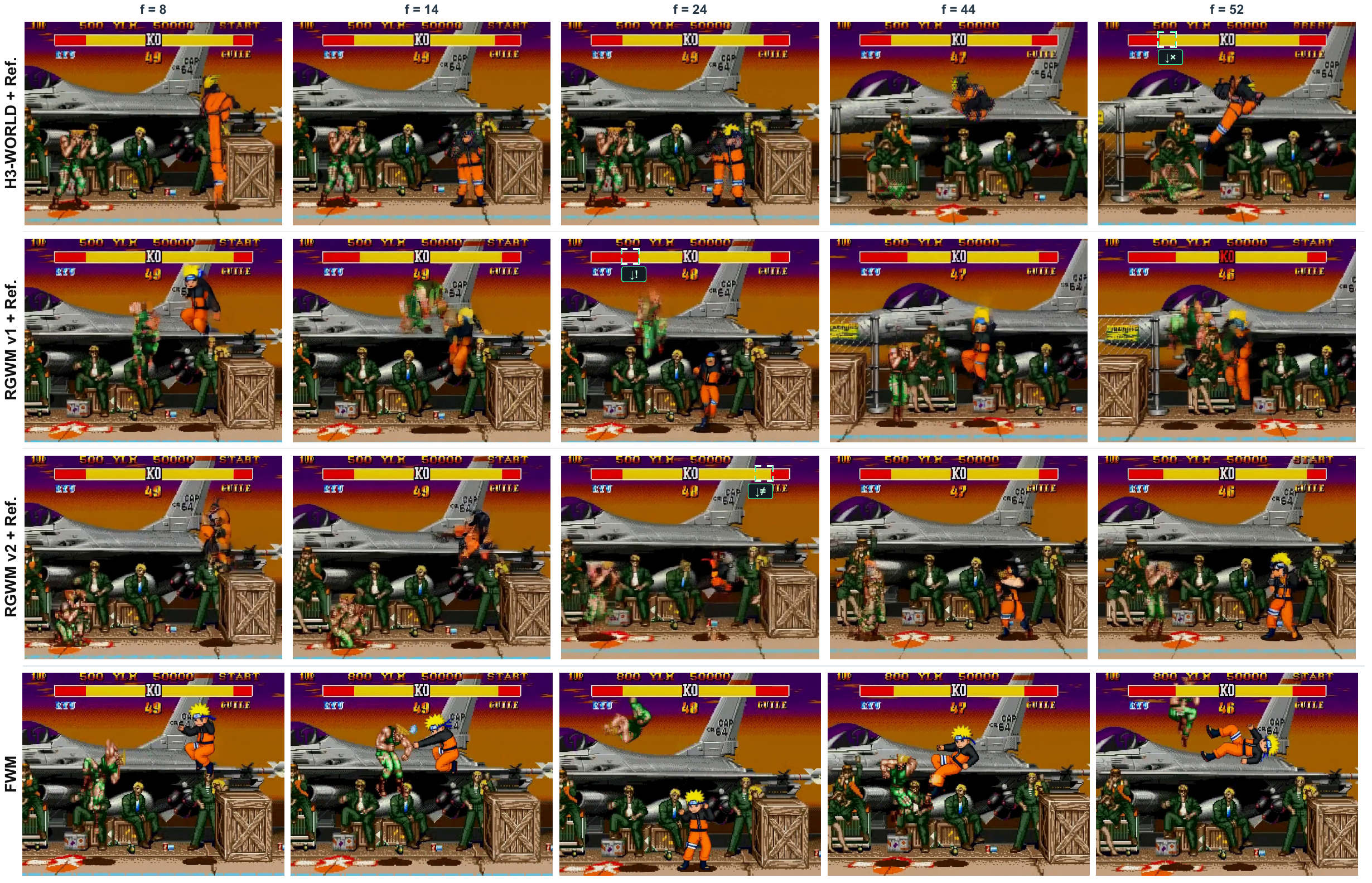}
  \caption{\textbf{Reference-Augmented Comparison (Naruto, Player, Side Switch).} Matched frames corresponding to row 2 of \cref{fig:appearance_ref}; HUD state icons follow \cref{fig:state_rep}.}
  \label{fig:app_naruto}
\end{figure*}

\begin{figure*}[p]
  \centering
  \includegraphics[width=\appendixfigwidth,height=0.42\textheight,keepaspectratio]{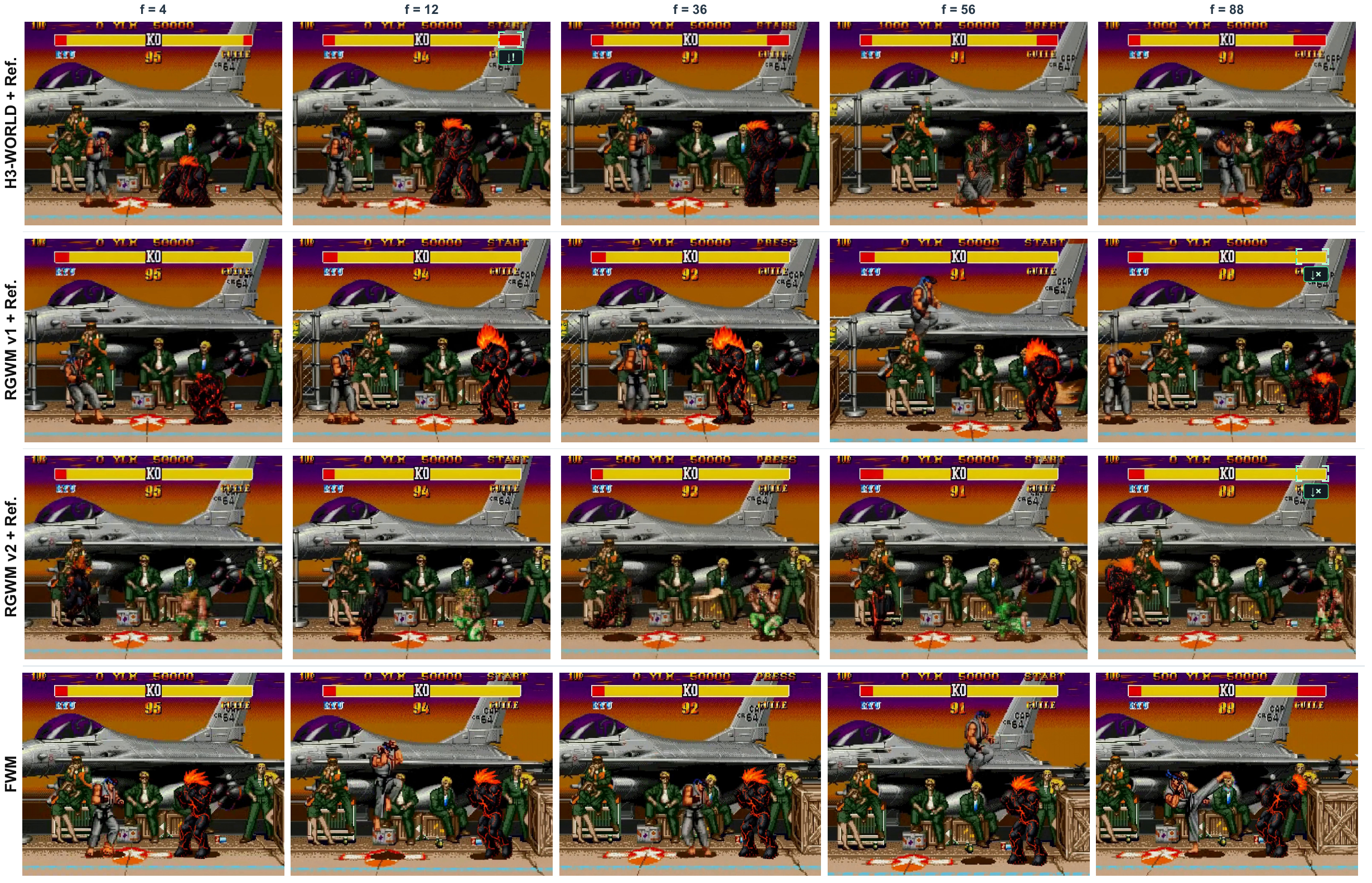}
  \caption{\textbf{Reference-Augmented Comparison (Lava Humanoid, NPC, No Side Switch).} Matched frames corresponding to row 3 of \cref{fig:appearance_ref}; HUD state icons follow \cref{fig:state_rep}.}
  \label{fig:app_lava}
  \includegraphics[width=\appendixfigwidth,height=0.42\textheight,keepaspectratio]{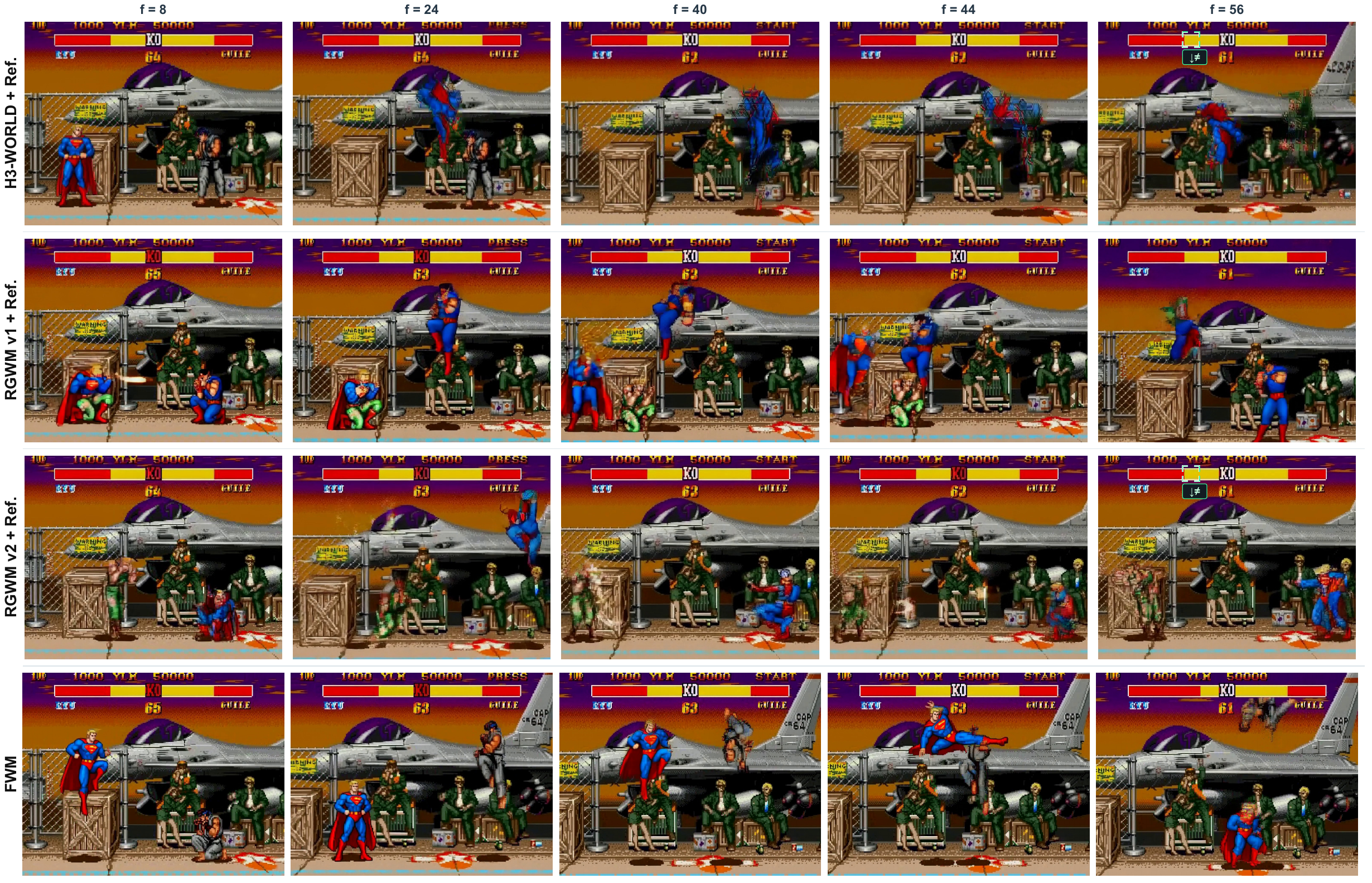}
  \caption{\textbf{Reference-Augmented Comparison (Superman, NPC, Side Switch).} Matched frames corresponding to row 4 of \cref{fig:appearance_ref}; HUD state icons follow \cref{fig:state_rep}.}
  \label{fig:app_superman}
\end{figure*}

We augment H3-WORLD~\cite{chen2026h3world}, ReactiveGWM v1~\cite{wang2026reactivegwm_v1}, and ReactiveGWM v2~\cite{wang2026reactivegwm_v2} with our reference-conditioning branch. Variants are evaluated on the same clips and frame indices as FWM, using explicit appearance references.
The comparisons in \cref{fig:app_saitama,fig:app_naruto,fig:app_lava,fig:app_superman} show character drift or garment blur during rapid contact and side switches, together with missed (\iconmiss{}), premature (\icondrop{}), or incomplete (\iconpart{}) health updates. FWM preserves appearance and the prescribed health updates more consistently in these examples.

\end{document}

%% file: preamble.tex
\usepackage[T1]{fontenc}
\usepackage{times}
\usepackage{graphicx,amsmath,amssymb,booktabs,array}
\usepackage[numbers,sort&compress]{natbib}
\usepackage{xcolor}
\usepackage{tikz}
\usetikzlibrary{arrows.meta,positioning,calc}
\definecolor{linkblue}{RGB}{36,89,137}
\definecolor{draftblue}{RGB}{51,91,133}
\usepackage{xurl}
\usepackage[breaklinks,colorlinks,allcolors=linkblue]{hyperref}
\usepackage[capitalize]{cleveref}
\definecolor{badgebg}{HTML}{121B22}
\definecolor{badgeroi}{HTML}{35EE9A}
\definecolor{badgetext}{HTML}{C6FFE0}
\definecolor{vokcol}{HTML}{168774}
\definecolor{vbadcol}{HTML}{D75C50}
\definecolor{vunkcol}{HTML}{6F7F8E}
\definecolor{vlinecol}{HTML}{D9E3E7}
\definecolor{ruleink}{HTML}{287B91}

\newcommand{\statebadge}[1]{%
  \tikz[baseline=-0.58ex]{%
    \node[fill=badgebg, draw=badgeroi, line width=0.45pt, rounded corners=1.2pt,
          inner xsep=2.2pt, inner ysep=1.1pt, text=badgetext,
          font=\fontsize{6.3}{6.3}\selectfont\sffamily\bfseries] {#1};%
  }%
}
\DeclareRobustCommand{\icondrop}{\statebadge{$\downarrow$!}}
\DeclareRobustCommand{\iconmiss}{\statebadge{$\downarrow\!\times$}}
\DeclareRobustCommand{\iconpart}{\statebadge{$\downarrow\!\neq$}}
\DeclareRobustCommand{\iconrise}{\statebadge{$\uparrow$!}}
\DeclareRobustCommand{\icondelay}{%
  \tikz[baseline=-0.58ex]{%
    \node[fill=badgebg, draw=badgeroi, line width=0.45pt, rounded corners=1.2pt,
          minimum width=11.5pt, minimum height=7.8pt, inner sep=0pt] (b) {};%
    \draw[badgetext, line width=0.55pt] (b.center) circle (2.3pt);%
    \draw[badgetext, line width=0.55pt] (b.center) -- ++(0,1.6pt) (b.center) -- ++(1.3pt,-0.5pt);%
  }%
}
\DeclareRobustCommand{\verdictok}{%
  \tikz[baseline=-0.55ex]{%
    \node[fill=white, draw=vlinecol, line width=0.45pt, rounded corners=1.2pt,
          minimum width=9.5pt, minimum height=7.8pt, inner sep=0pt] (v) {};%
    \draw[vokcol, line width=0.85pt] ([ shift={(-2.3pt,-0.1pt)} ]v.center) -- ++(1.3pt,-1.5pt) -- ++(3.0pt,3.3pt);%
  }%
}
\DeclareRobustCommand{\verdictbad}{%
  \tikz[baseline=-0.55ex]{%
    \node[fill=white, draw=vlinecol, line width=0.45pt, rounded corners=1.2pt,
          minimum width=9.5pt, minimum height=7.8pt, inner sep=0pt] (v) {};%
    \draw[vbadcol, line width=0.8pt] ([ shift={(-2.0pt,-2.0pt)} ]v.center) -- ++(4.0pt,4.0pt) ([ shift={(-2.0pt,2.0pt)} ]v.center) -- ++(4.0pt,-4.0pt);%
  }%
}
\DeclareRobustCommand{\verdictunk}{%
  \tikz[baseline=-0.55ex]{%
    \node[fill=white, draw=vlinecol, line width=0.45pt, rounded corners=1.2pt,
          inner xsep=2.2pt, inner ysep=0.9pt, text=vunkcol,
          font=\fontsize{6.3}{6.3}\selectfont\sffamily\bfseries] {?};%
  }%
}

\newcommand{\resultfigwidth}{\textwidth}
\newcommand{\tallresultfigwidth}{\textwidth}
\newcommand{\appendixfigwidth}{\textwidth}
\newcommand{\referencegalleryfigwidth}{\textwidth}

%% file: tables/04_reference_binding.tex
\begin{table}[!t]
\centering
\small
\setlength{\tabcolsep}{5pt}
\caption{\textbf{Effect of Identity Binding.}
Variants share reference inputs and training budgets, differing only in identity binding.
Adherence is a percentage.
Values are ranked within each test subset.
}
\label{tab:binding_control}
\begin{tabular}{@{}lcrr@{}}
\toprule
Test Subset
& Binding
& \shortstack{Adherence $\uparrow$}
& \shortstack{DINO Similarity $\uparrow$} \\
\midrule
Non-Switch & Off & \underline{78.43} & \underline{0.47} \\
Non-Switch & On  & \textbf{99.16} & \textbf{0.61} \\
\midrule
Side-Switch & Off & \underline{53.68} & \underline{0.35} \\
Side-Switch & On  & \textbf{98.27} & \textbf{0.60} \\
\bottomrule
\end{tabular}
\end{table}

%% file: tables/01_control_representations.tex
\begin{table*}[!t]
\centering
\small
\setlength{\tabcolsep}{4pt}
\caption{\textbf{Interaction Fidelity across Control Representations.}
We compare variants with different conditioning signals to assess the effects of interaction information and its encoding.
Rates are percentages, and Health MAE is in percentage points of maximum health.
Across tables, bold and underlined values mark the best and second-best results.
}
\label{tab:main}
\begin{tabular}{@{}lcrrr@{}}
\toprule
Conditioning &
\shortstack{Action $\uparrow$\\Move / Attack / Skill} &
\shortstack{Reaction $\uparrow$} &
\shortstack{Health MAE $\downarrow$} &
$J\uparrow$ \\
\midrule
Actions Only
& 70.27 / 56.22 / 58.64  & 74.43  & 3.42 & 34.18 \\
Actions + Rule Text
& 71.71 / 68.40 / 57.35 & 80.16  & 3.07 & 46.37 \\
Spatial Proxy Only
& \underline{99.19} / 92.27 / 46.58 & 78.28 & 2.31 & 33.46 \\
\midrule
Spatial Proxy + State Tokens
& 94.48 / 91.23 / 82.42 & 87.17 & \underline{1.23} & 63.28 \\
Spatial Proxy + Text
& 96.72 / \underline{93.63} / \underline{92.43} & \underline{92.24} & 2.78 & \underline{76.35} \\
FWM Interaction Proxy
& \textbf{99.78} / \textbf{99.24} / \textbf{99.06} & \textbf{99.79} & \textbf{0.15} & \textbf{98.48} \\
\bottomrule
\end{tabular}
\end{table*}

%% file: tables/02_existing_methods.tex
\begin{table*}[!t]
\centering
\small
\setlength{\tabcolsep}{4pt}
\caption{\textbf{Comparison with Adapted Game World Models.}
Baselines are adapted to FightWorld Corpus.
+Ref denotes the addition of our reference-conditioning branch.
Rates are percentages, and Health MAE is in percentage points of maximum health.
DINO measures cosine similarity.
A dash indicates an inapplicable metric.
}
\label{tab:method_comparison}
\begin{tabular}{@{}lcrrcrr@{}}
\toprule
Method
& \shortstack{Action $\uparrow$\\Move / Attack / Skill}
& Reaction $\uparrow$
& \shortstack{Health\\MAE $\downarrow$}
& \shortstack{Adherence $\uparrow$\\Character / Scene}
& \shortstack{Character\\DINO $\uparrow$} & \shortstack{Scene\\DINO $\uparrow$} \\
\midrule
\multicolumn{7}{@{}l}{
\emph{Without Reference-Image Conditioning}} \\
H3-WORLD
& 43.85 / 62.61 / 47.58
& 48.25 & 8.75 & \textemdash{} & \textemdash{} & \textemdash{} \\
ReactiveGWM v1
& 70.15 / 78.98 / 48.55
& 35.56 & 4.59 & \textemdash{} & \textemdash{} & \textemdash{} \\
ReactiveGWM v2
& 50.24 / 59.70 / 16.67
& 49.21 & 4.65 & \textemdash{} & \textemdash{} & \textemdash{} \\
\midrule
\multicolumn{7}{@{}l}{
\emph{With Reference-Image Conditioning}} \\
H3-WORLD + Ref.
& 36.91 / 62.70 / 42.03
& 42.22 & 7.57 & 88.10 / \textbf{100.00} & 0.47 & \underline{0.68} \\
ReactiveGWM v1 + Ref.
& 63.16 / \underline{85.53} / \underline{53.86}
& 33.97 & 4.79 & 89.29 / \textbf{100.00} & \underline{0.50} & 0.64 \\
ReactiveGWM v2 + Ref.
& \underline{72.22} / 74.10 / 34.78
& \underline{57.14} & \underline{4.04} & \underline{91.67} / \textbf{100.00} & 0.41 & 0.65 \\
\midrule
\textbf{FWM (Ours)}
& \textbf{99.78} / \textbf{99.24} / \textbf{99.06}
& \textbf{99.79}
& \textbf{0.15}
& \textbf{99.12} / \textbf{100.00}
& \textbf{0.60} & \textbf{0.72} \\
\bottomrule
\end{tabular}
\end{table*}

%% file: tables/05_proxy_ablations.tex
\begin{table*}[!t]
\centering
\small
\setlength{\tabcolsep}{6pt}
\caption{\textbf{Contributions of Proxy Cues.}
Each ablated variant omits one cue family and is retrained under the same training setting.
Rates are percentages, and Health MAE is in percentage points of maximum health.
}
\label{tab:ablation}
\begin{tabular}{@{}lcrrr@{}}
\toprule
Variant
& \shortstack{Action $\uparrow$\\Move / Attack / Skill}
& Reaction $\uparrow$
& \shortstack{Health\\MAE $\downarrow$}
& $J$ $\uparrow$ \\
\midrule
\textbf{FWM}
& \underline{99.78} / \textbf{99.24} / \underline{99.06}
& \textbf{99.79} 
& \textbf{0.15}
& \textbf{98.48} \\
\midrule
w/o Command Marks
& 99.07 / 86.42 / 87.87
& 97.93
& 0.28
& \underline{85.26} \\
w/o Damage-Event Cues
& 99.16 / 98.92 / 98.91
& 80.42
& 0.27
& 78.38 \\
w/o Health Bars
& \textbf{99.84} / \underline{99.08} / \textbf{99.12}
& \underline{98.31}
& 2.18
& 57.24 \\
w/o Skill Event Trigger
& 99.21 / 99.03 / 79.28
& 98.07
& \underline{0.26}
& 78.81 \\
\bottomrule
\end{tabular}
\end{table*}

%% file: tables/03_rule_interventions.tex
\begin{table}[!t]
\centering
\small
\setlength{\tabcolsep}{5pt}
\caption{\textbf{Interaction Fidelity under Rule Changes.}
Within each rule setting, encodings share executed records and are ranked separately.
Default rules use original damage without immunity or lifesteal.
Paired values denote reduced / amplified damage.
Health MAE is in percentage points of maximum health, and $J_{\mathrm{rule}}$ is a percentage.
}
\label{tab:rules}
\begin{tabular}{@{}lcc@{}}
\toprule
Encoding &
\shortstack{Health MAE $\downarrow$} &
$J_{\mathrm{rule}}\uparrow$ \\
\midrule
\multicolumn{3}{@{}l}{\emph{Default Rules}} \\
Spatial Proxy + State Tokens & \underline{1.87} & \underline{89.26} \\
Spatial Proxy + Text      & 2.42 & 76.38 \\
FWM               & \textbf{0.16} & \textbf{100.00} \\
\midrule
\multicolumn{3}{@{}l}{\emph{Damage Magnitude}} \\
Spatial Proxy + State Tokens
& \underline{1.63} / \underline{2.48} & \underline{74.62} / \underline{65.37} \\
Spatial Proxy + Text
& 3.17 / 4.86 & 56.28 / 43.71 \\
FWM
& \textbf{0.18} / \textbf{0.22} & \textbf{97.76} / \textbf{98.84} \\
\midrule
\multicolumn{3}{@{}l}{\emph{Limited-Hit Immunity}} \\
Spatial Proxy + State Tokens & \underline{4.28} & \underline{49.36} \\
Spatial Proxy + Text      & 8.73 & 24.18 \\
FWM               & \textbf{0.21} & \textbf{99.92} \\
\midrule
\multicolumn{3}{@{}l}{\emph{Lifesteal}} \\
Spatial Proxy + State Tokens & \underline{5.67}  & \underline{37.42}  \\
Spatial Proxy + Text      & 10.46  & 16.83  \\
FWM               & \textbf{0.19}  & \textbf{99.87} \\
\bottomrule
\end{tabular}
\end{table}